\documentclass{extarticle}
\usepackage{microtype}
\usepackage{graphicx}
\usepackage{subfigure}
\usepackage{booktabs} 
\usepackage{lipsum}
\usepackage{multicol}
\usepackage{bm}
\usepackage{float}
\usepackage{multirow}
\usepackage{listings}
\usepackage{array}
\usepackage{listings}
\usepackage{xcolor}

\usepackage{hyperref}

\usepackage[accepted]{icml2025}

\usepackage{amsmath}
\usepackage{amssymb}
\usepackage{mathtools}
\usepackage{amsthm}

\usepackage[capitalize,noabbrev]{cleveref}

\theoremstyle{plain}

\theoremstyle{definition}

\theoremstyle{remark}

\usepackage[textsize=tiny]{todonotes}

\icmltitlerunning{Memorization and Knowledge Injection in Gated LLMs}

\begin{document}

\twocolumn[
\icmltitle{Memorization and Knowledge Injection in Gated LLMs}



\icmlsetsymbol{equal}{*}

\begin{icmlauthorlist}
\icmlauthor{Xu Pan}{equal,yyy}
\icmlauthor{Ely Hahami}{equal,xxx,kempner}
\icmlauthor{Zechen Zhang}{yyy,xxx}
\icmlauthor{Haim Sompolinsky}{yyy,kempner,zzz}
\end{icmlauthorlist}

\icmlaffiliation{xxx}{Harvard University, Cambridge, MA}
\icmlaffiliation{yyy}{Center for Brain Science, Harvard University, Cambridge, MA}
\icmlaffiliation{zzz}{Edmond and Lily Safra Center for Brain Sciences, Hebrew University, Jerusalem, Israel}

\icmlaffiliation{kempner}{Kempner Institute, Harvard University, Cambridge, MA}


\icmlcorrespondingauthor{Haim Sompolinsky}{hsompolinsky@mcb.harvard.edu}

\icmlkeywords{Machine Learning, ICML}

\vskip 0.3in
]



\printAffiliationsAndNotice{\icmlEqualContribution}

\begin{abstract}

Large Language Models (LLMs) currently struggle to sequentially add new memories and integrate new knowledge. These limitations contrast with the human ability to continuously learn from new experiences and acquire knowledge throughout life. Most existing approaches add memories either through large context windows or external memory buffers (e.g., Retrieval-Augmented Generation), and studies on knowledge injection rarely test scenarios resembling everyday life events. In this work, we introduce a continual learning framework, Memory Embedded in Gated LLMs (MEGa), which injects event memories directly into the weights of LLMs. Each memory is stored in a dedicated set of gated low-rank weights. During inference, a gating mechanism activates relevant memory weights by matching query embeddings to stored memory embeddings. This enables the model to both recall entire memories and answer related questions. On two datasets - fictional characters and Wikipedia events - MEGa outperforms baseline approaches in mitigating catastrophic forgetting. Our model draws inspiration from the complementary memory system of the human brain.

\end{abstract}

\section{Introduction}

The rapid advancement of large language models (LLMs) has dramatically reshaped our understanding of the capabilities and potential of AI systems. These models have proven invaluable across diverse fields, assisting researchers and practitioners. For cognitive neuroscientists, LLMs offer an unprecedented opportunity to study an intelligent system that, although not human, can process natural language and advanced cognitive functions, and compare it to human cognition as well as its underlying brain mechanisms \cite{hagendorff2023machine, binz2024turning, coda-forno2024cogbench}. Such comparisons advance human cognitive neuroscience and may lead to more powerful AI systems \cite{silver_sutton_2025_experience}. In this paper, we propose to study long-term declarative memory (e.g. episodic memory and semantic memory), one of the hallmarks of human cognition, using LLMs as a model cognitive system. To achieve this, we augment a pretrained LLM with gated memory modules, enabling rapid continual encoding and retrieval of memories while mitigating catastrophic forgetting.

Classical models of long-term memory in neural networks are based on the paradigm of associative memory in recurrent neural networks (RNNs), such as the Hopfield model \cite{hopfield1982neural}, where each memory corresponds to a stable activation pattern of the network. This paradigm was later extended to the memorization of sequences of states \cite{kleinfeld1988associative, kanter1987associative}. A common feature of these models is the use of Hebbian-like learning rules, which are inherently incremental and align with the continual nature of long-term memory. However, Hebbian learning is severely limited in that it can only store random, uncorrelated patterns, with accurate recall possible only when the number of memories does not exceed a capacity limit that scales linearly with the network size. Beyond this limit, the system suffers from catastrophic forgetting (CF) \cite{french1999catastrophic}. When the memorized states are correlated, the capacity is drastically reduced to just a few memories—regardless of network size—due to strong interference between them \cite{lowe1998storage}. Some learning rules partially address this limitation but require batch learning of all memories \cite{kanter1987associative, gardner1988optimal}, which is unsuitable for modeling continual learning (CL) of memories. Moreover, even these batch models of attractor networks degrade significantly when the stored patterns are analog, rather than binary \cite{schonsberg2021efficiency}. The challenge of mitigating CF in CL, especially when dealing with realistic, highly correlated analog data, remains a persistent obstacle in both machine learning and cognitive neuroscience. Most current regularization approaches to CL perform poorly when faced with sequences of correlated tasks \cite{shan2024order}. Rehearsal-based methods demand substantial memory resources, and existing models of spontaneous memory reactivation in Hebbian RNNs \cite{shaham2022stochastic} offer limited effectiveness when the memories are correlated.

Another fundamental limitation of classical memory models in “shallow” RNNs is their tendency to encode memories as isolated knowledge items. In contrast, real-world memories are composed of events rich in semantic structure, with elements that are typically already familiar to the organism. As a result, new factual memories must be embedded within or interact closely with an existing, fully developed semantic system. The same principle applies to working memory, where ongoing experiences must integrate with extensive semantic knowledge to support current perceptual or motor tasks. Most neural network models of working memory, however, store items in transient activity patterns that are independent of semantic content \cite{hochreiter1997long}. Although several cognitive theories have proposed memory systems that use pointers or binding mechanisms to associate memory items with their context \cite{norris2017short, blouw2016concepts}, no existing model addresses these fundamental challenges at a scale that matches human memory. Finally, a key function of attractor models is pattern completion, achieved through the convergence of all states within a memory’s basin of attraction to a fixed point—a property known as content-addressable memory. Yet, this concept requires substantial revision, as the recall of semantically rich memories typically does not begin from randomly corrupted inputs.

Several studies have explored memory functions in large language models (LLMs), both for AI applications and as analogies to human memory systems \cite{raccah2022memory, wang2024towards, janik2023aspects, gutierrez2024hipporag, fountas2024human, gershman2025key, silver_sutton_2025_experience}. Within LLMs, three primary analogies to memory have emerged: (1) tokens in the context window, (2) an external database connected to the model, and (3) the model’s internal weights. Each of these representations reflects certain aspects of human memory, yet all fall short of fully modeling human long-term memory.

In-context learning \cite{brown2020language} appears to avoid the limitations of classical associative memory models: it does not suffer from catastrophic forgetting and can learn new memories, even when they are correlated with existing ones, by smoothly integrating them into the LLM’s existing semantic knowledge. However, the demands of long-term memory may exceed the capacity of context windows \cite{bai2023longbench}. Notably, the context window more closely resembles human working memory. Attempting to unify working and long-term memory into a single representation is a biologically implausible model of memory, as these two functions rely on distinct cognitive resources and brain systems \cite{fountas2024human}.

Retrieving memories from an external database into the LLM’s context window is known as Retrieval-Augmented Generation (RAG). RAG has proven to be an effective method for enhancing the memory capabilities of LLMs and is widely used in real-world applications \cite{lewis2020retrieval}. In this approach, user queries are processed by an embedding model to retrieve relevant entries—based on a similarity metric—from the database. These entries are then inserted into the LLM’s context window, enabling the model to generate informed responses by combining its inherent capabilities with the retrieved knowledge.

Studies have also investigated RAG in continual learning settings, where external databases are incrementally updated over time while preserving knowledge consistency \cite{li2024language, gupta2024stackfeed, modarressi2024memllm, zhang2025dh, fan2025research}. Variants of RAG have also been developed that store and retrieve key-value pairs (i.e., KV-cache) or other compressed representations of knowledge, rather than raw text \cite{modarressi2024memllm, das2024larimar, xiao2024infllm, yang2024memory3, qian2025memorag, chan2024don}.

Despite RAG's usefulness in applications, storing memories in an external text-based database is not an attractive candidate for modeling human long-term memory, since this approach more accurately resembles an external environment with which a subject interacts. Rather, injecting memories by fine-tuning the LLM parameters more closely mirrors the biological process where long-lasting synaptic changes of the neural circuits sustain long-term memory. However, fine-tuning LLMs for long-term memory faces several challenges, such as catastrophic forgetting \cite{luo2023empirical, chen2023chatgpt, zhai2024investigating, gupta2024model, zhang2024dissecting, zhu2024model, yang2024butterfly, song2025complete}, lack of generalization \cite{berglund2023reversal, ovadia2023fine, yang2024unveiling}, slow learning and hallucination \cite{gekhman2024does, kang2024unfamiliar}.

To address these challenges, we introduce MEGa (Memory Embedded in Gated LLMs), a long-term memory framework designed to enable LLMs to sequentially store new memories in a manner that reflects key aspects of human memory.

To ensure biological plausibility, MEGa encodes new memories by fine-tuning the network’s weights. To mitigate catastrophic forgetting, it employs a gating mechanism that, at inference time, routes input queries to a collection of gated memory modules and activates those most relevant to the query.

We show that MEGa is capable not only of retrieving the learned memories but also of performing question-answering (QA) tasks based on them, demonstrating the successful integration of the memories into the knowledge base of the LLM.

Across two datasets—Fictional Character and Wikipedia 2024 Events—and two tasks—memory recall and question answering—MEGa outperforms baseline continual learning (CL) techniques in mitigating both the forgetting of newly acquired memories and the degradation of general language capabilities. These results suggest that MEGa is a promising model for capturing certain aspects of human long-term memory and the structure of underlying brain circuits.

\section{Related Works}

\subsection{Continual Learning and Catastrophic Forgetting}

Several studies have shown that LLMs suffer from catastrophic forgetting of previously acquired knowledge and skills during continual learning (CL) on new tasks \cite{luo2023empirical, chen2023chatgpt, zhai2024investigating, song2025complete, zhang2024dissecting, zhu2024model}. To address this issue, various CL methodologies have been adapted for use with LLMs, including regularization-based approaches \cite{lee2019mixout, zhang2020revisiting, chen2020recall, kotha2023understanding, zhu2024model}, rehearsal-based strategies \cite{sun2019lamol, xu2024magpie, huang2024mitigating}, and architecture-based techniques \cite{hartvigsen2024aging}. Our proposed method MEGa can be considered an architecture-based method, as it adds new components and gating to an existing model.

\subsection{Knowledge Injection}

Injecting new knowledge into pretrained LLMs has recently garnered significant attention \cite{hsueh2024editing, shi2024continual, zhang2024comprehensive, thede2025understanding}. A straightforward approach involves fine-tuning the model on the knowledge text \cite{ovadia2023fine, gangadhar2024model}, or on the answers when the knowledge is provided in the form of QA pairs \cite{mecklenburg2024injecting}. More recent methods aim to localize weight updates by identifying a knowledge-relevant subspace of the model’s weights \cite{meng2022locating, mitchell2021fast}, or by distilling knowledge from the context window into the model’s parameters \cite{qi2024context, padmanabhan2024propagating, wangself, kujanpaa2024knowledge}. However, there is evidence that these approaches are not significantly more effective than standard fine-tuning \cite{gangadhar2024model, thede2025understanding}.

However, many of these methods—and their associated experimental setups—fall short of emulating how humans acquire long-term episodic memories. Most widely used knowledge editing datasets \cite{levy2017zero, meng2022locating, zhang2024comprehensive, thede2025understanding} represent knowledge as simple subject-relation-object triples. This format, however, lacks the richness and complexity of human episodic memory, which typically involves detailed semantic representations of personal experiences. As a result, methods developed using subject-relation-object datasets \cite{meng2022locating, meng2022mass, mitchell2021fast, wang2024wise, qi2024context, yu2024melo} are often not directly applicable for injecting knowledge expressed in paragraph-level event descriptions. Another limitation is that success in these methods is usually defined as the model generating the correct object given a subject-relation pair—whereas human episodic memory is far more flexible: a single memory can be triggered by a variety of cues and used to answer diverse questions.

\subsection{Gating Networks}

Our model, MEGa, uses gating units to route queries to the most relevant stored memories. In general, gating networks function by selectively activating or suppressing connection paths based on the context or input provided to the system. Both empirical studies \cite{hochreiter1997long, chung2014empirical, sezener2021rapid, veness2021gated} and theoretical analyses \cite{saxe2022neural, li2022globally} have shown that gated architectures are effective at mitigating catastrophic forgetting and are well-suited for training across multiple tasks. Gating mechanisms are widely used in modern deep neural networks. One prominent example is the Mixture of Experts (MoEs) architecture, a type of gated network that has gained popularity and contributes to some of the state-of-the-art LLMs \cite{shazeer2017outrageously, fedus2021switch, jiang2024mixtral}.

Low-Rank Adaptation (LoRA) is a popular parameter-efficient fine-tuning technique, which freezes a pretrained weight matrix $W_{\text{PT}}$ and injects trainable rank decomposition matrices into each layer of the Transformer architecture, such that \(W_{\text{FT}} := W_{\text{PT}} +AB\) \cite{hu2021lora}. Here, $W_{\text{FT}} \in \mathbb{R}^{d \times k}$ denotes the fine-tuned weight matrix, \(A \in \mathbb{R}^{d\times r}\) and \(B \in \mathbb{R}^{r\times k}\) are trainable update matrices, and \(d\) and $k$ are input and output dimensions.

Several studies have explored using gated LoRA modules to enhance fine-tuning performance across multiple tasks. Some approaches determine per-token gating weights based on routing networks \cite{jung2024pmoe, buehler2024x, luo2024moelora, xu2024meteora, zhao2024loraretriever}, while others rely on local activation levels \cite{wang2024wise, zhu2024initializing}. In contrast, our work determines gating weights based on the semantic similarity between the query and the stored knowledge, which serves as a global signal for all the layers and tokens.  The MELO framework \cite{yu2024melo} is the most comparable to MEGa in this respect, although our datasets and evaluations are more complex and better capture the characteristics of human long-term episodic memory.

\begin{figure*}[ht]
\begin{center}
\centerline{\includegraphics[width=1.5\columnwidth]{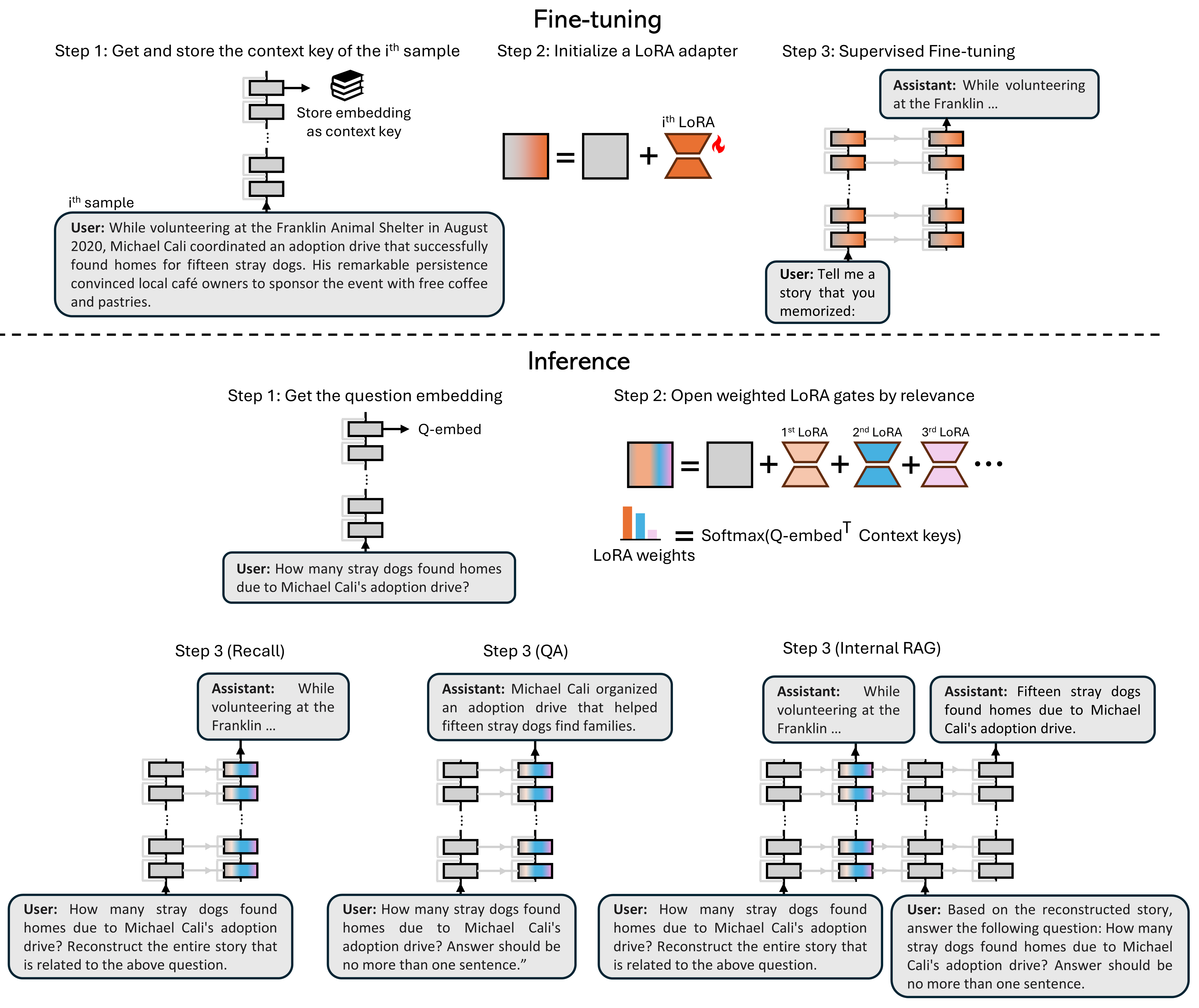}}
\caption{Illustration of our fine-tuning and inference pipeline. We store the embedding of each sample as its context key and initialize a LoRA adapter for each sample (i.e., raw text paragraphs). LoRA adapters are fine-tuned individually using a constructed user prompt. During inference, the base model processes the query to generate its embedding, which is then used to compute the LoRA gating weights. Questions can be augmented with additional instructions depending on the task. For the recall task, the model is prompted to reconstruct the entire relevant story. For the QA task, the model directly answers the question, with the instruction “Answer should be no more than one sentence” appended to ensure concise responses. In the internal RAG (iRAG) setup, the model first reconstructs the relevant story and then answers the question based on the reconstructed story. All inputs are formatted according to the Llama-3 instruct format.}
\label{Illustration}
\end{center}
\end{figure*}

\section{Methods}

\subsection{Problem Formulation}
\label{3.1}

Our goal is to build a system capable of memorizing multiple memories that arrive sequentially, and later retrieving those memories and extracting knowledge from them—while minimizing destructive interference both among the memories themselves and across the model’s general language abilities. This mirrors how the human brain continuously acquires declarative memory through experience. Individual memories are rich in semantic context, such as events involving people, actions, time, and locations. 

The datasets used in this work consist of a set $\{D_1, D_2, ... D_n\}$ where each sample, $D_i$, is a short paragraph describing an event that the model has not seen before. These memories are learned by a pretrained LLM through fine-tuning. Importantly, rather than being presented in a batch, the samples are provided to the model sequentially and exclusively—one at a time—reflecting both the nature of human experience and real-world scenarios where task data arrives incrementally. The goal is for the model, after fine-tuning, to (1) reconstruct each individual sample when given an relevant cue, and (2) answer questions related to one or more of the fine-tuned texts. We refer to the first as the recall task and the second as the QA task, corresponding to the model's memory and knowledge, respectively.

\begin{table*}[ht]
\vskip 0.15in
\begin{center}
\begin{sc}
\begin{tabular}{p{0.11\linewidth} | >{\scriptsize}p{0.078\linewidth} >{\scriptsize}p{0.078\linewidth} >{\scriptsize}p{0.07\linewidth} >{\scriptsize}p{0.07\linewidth} | >{\scriptsize}p{0.078\linewidth} >{\scriptsize}p{0.078\linewidth} >{\scriptsize}p{0.07\linewidth} >{\scriptsize}p{0.07\linewidth}}
\toprule
& \multicolumn{4}{c|}{Fictional character} & \multicolumn{4}{c}{Wiki events} \\
&QA Acc \% $\uparrow$&Log prob $\downarrow$&Recall cos $\uparrow$&MMLU acc \% $\uparrow$&QA Acc \% $\uparrow$& Log prob $\downarrow$& Recall cos $\uparrow$& MMLU acc \%$\uparrow$\\
\midrule

Base & 0.13±0.27 & -4.85±0.39 & 0.587±0.024 & \textbf{62.56} & 8.17±2.38 & -3.25±0.14 & 0.665±0.018 & \textbf{62.56} \\
RAG & \textbf{82.57±6.37} & -2.57±0.46 & 0.881±0.002 & \textbf{62.56} & \textbf{88.83±2.71} & \textbf{-1.76±0.15} & 0.889±0.006 & \textbf{62.56} \\
Full & 12.60±11.03 & -2.92±0.30 & 0.554±0.066 & 55.65±1.54 & 17.90±9.57 & -2.00±0.28 & 0.526±0.149 & 56.28±1.03 \\
Full L2& 25.47±10.32 & -3.14±0.45 & 0.609±0.084 & 55.25±1.79 & 31.33±11.69 & -2.07±0.33 & 0.623±0.163 & 55.98±1.48 \\
Full ewc& 14.10±8.08 & -2.82±0.29 & 0.542±0.064 & 55.24±1.75 & 21.43±13.47 & -2.02±0.32 & 0.544±0.149 & 56.21±0.98 \\
Full batch& 54.77±3.27 & -2.77±0.28 & 0.853±0.013 & 60.62±1.05 & 76.87±5.72 & -1.87±0.26 & \textbf{0.926±0.011} & 60.90±1.34 \\
Lora& 0.80±1.20 & -4.13±0.44 & 0.485±0.038 & 47.94±2.11 & 0.53±0.72 & -3.02±0.49 & 0.243±0.055 & 46.88±1.42 \\
Lora L2& 12.27±6.98 & -3.30±0.42 & 0.528±0.058 & 47.49±2.53 & 18.20±9.81 & -2.35±0.41 & 0.454±0.144 & 49.12±2.30 \\
MEGa {\scriptsize (ours)}& 72.53±6.79 & \textbf{-2.12±0.25} & \textbf{0.901±0.011} & 61.75±0.53 & 78.03±3.50 & -1.86±0.18 & 0.921±0.013 & 61.99±0.57 \\
iRAG {\scriptsize (ours)}& 80.67±6.09 & -2.84±0.33 & \textbf{0.901±0.011} & 61.75±0.53 & 84.70±4.37 & -2.24±0.21 & 0.921±0.013 & 61.99±0.57 \\
\bottomrule
\end{tabular}
\end{sc}
\end{center}
\caption{Evaluation after continual learning on 50 samples. QA accuracy (evaluated by a GPT judge) and log probability assess the model’s question-answering ability, while recall cosine similarity measures memorization performance. MMLU accuracy reflects the degree of catastrophic forgetting in the model’s general knowledge. Results are reported as the mean and standard deviation across 20 dataset partitions. For RAG in the reconstruction task, we report the metric:
\(
\text{hit rate} \cdot 1 + (1 - \text{hit rate})\cdot c, 
\)
where $c$ is the average cosine similarity between randomly selected training samples, and the hit rate is the percentage of times RAG includes the correct sample in the LLM’s context window. The MMLU accuracy of RAG is reported same as the base model.}
\vskip -0.1in
\label{maintable}
\end{table*}

\subsection{Model and Datasets}

All experiments are conducted using the Llama-3.1-8B-Instruct model, referred to as the “base” model. We fine-tune it on two datasets:

\textbf{Fictional Character Dataset}: We generate synthetic data by prompting GPT-4.5 (see Appendix \ref{section:prompts}) to produce 50 paragraphs (i.e., samples) based on a manually created template. Each paragraph describes a specific event in the life of a fictional character with a randomly generated name. To align with our goal of modeling episodic memory, the paragraphs capture concrete events (e.g., a basketball game, a trip to the Swiss Alps). This procedure is repeated 20 times with different random names, resulting in 20 dataset partitions. Continual learning experiments are performed separately on each partition, and unless otherwise noted, reported evaluation metrics include the mean and standard deviation across the 20 partitions. The average length of each story is 41.93 words. Following prior work \cite{mecklenburg2024injecting, mosbach2023few, ovadia2023fine}, we also generate nine paraphrases for each sample during fine-tuning using GPT-4.5.

Lastly, to construct the evaluation set, we prompt GPT-4.5 to generate three QA pairs for each sample, based on its content (see Appendix \ref{section:prompts} for prompt details and Appendix \ref{character_dataset_example} for an example).

\textbf{Wikipedia 2024 Events}: Following previous knowledge-injection studies \cite{mecklenburg2024injecting, ovadia2023fine, zhang2024comprehensive}, we use Wikipedia articles on recent events as our second dataset. Since the knowledge cutoff date for Llama-3.1-8B-Instruct is December 2023, we crawled all Wikipedia articles categorized under “2024\_by\_month” and its subcategories. We further filtered out articles that were first created before 2024. For fine-tuning, we use the first section of each retained article, which typically provides a summary of the event. From these, we randomly sampled 1,000 articles with character counts between 200 and 300 (average word count: 41.55) to form our fine-tuning dataset. As with the Fictional Character dataset, we partition these samples into 20 subsets for continual learning experiments and generate corresponding paraphrases and QA pairs (see Appendix \ref{wiki_dataset_example} for an example).

\textbf{Compositional questions}: In addition to QA pairs based on individual samples, we prompt GPT-o3-mini (see full prompt in Appendix \ref{compquestions}) to generate an evaluation set of 500 compositional questions—each requiring knowledge from exactly two distinct samples to answer correctly (see Appendix \ref{Appendix-E.1} for an example).

\subsection{Memory Embedded in Gated LLMs (MEGa)}

The fine-tuning and inference algorithm of MEGa is shown schematically in Figure \ref{Illustration}.

\subsubsection{Fine-tuning}

During fine-tuning on a new sample, $D_i$, we first get and store the embedding of this sample as ``context key", \(K_i = f(D_i)\), where $f$ is an embedding function that maps a sequence of tokens into a fixed-sized vector containing semantic-rich information about the text. We employ embeddings generated internally by the base model. Specifically, we define $f$ as the average of the internal activations computed near the end of the base model—precisely, the input to the final MLP layer \cite{muennighoff2022sgpt}. We explore various embedding layers and embedding models in Appendix \ref{section:Supplementary}.

Next, we fine-tune a set of weights denoted as LoRA adapter. Each LoRA adapter consists of a set of low-rank matrices, \(\{A^l,B^l\}\), applied on specified layers \(\{l\}\) and modules. For each sample, our gated-LoRA method initializes and trains exactly one LoRA adapter. Specifically, when training on the $i$th sample $D_i$, \(\{A_i^l,B_i^l\}\) are (see the Appendix \ref{section:Fine-Tuning Settings} on initialization) added to the pre-trained weights $W^l_{PT}$ to form the new set of weights $\Theta_{i} $:

\begin{equation}
\text{\textit{fine-tuning weights:}}\:\: \Theta_{i} =  \{W^l_{PT} + A^l_iB^l_i\}_ .
\end{equation}
The LoRA adapter weights are fine-tuned by minimizing the following loss,  

\begin{equation}
L_i(\{A^l_i,B^l_i\}) = - \log p_{\Theta_{i}}(D_i \mid x),
\end{equation}

Here, $x$ refers to an appropriate \textit{fine-tuning prompt}. Common choices include the “begin of text” token (i.e., continued pretraining) and formatted user queries (i.e., supervised fine-tuning). We found that using the “begin of text” token resulted in poor performance on the Fictional Character dataset (see Appendix \ref{section:superpervised_finetuning}). For our main experiments, we use the user query prompt “Tell me a story that you memorized.” (see Figure \ref{Illustration} and Appendix \ref{section:superpervised_finetuning}). Other prompt formats are also possible (see Appendix \ref{section:superpervised_finetuning} for details).

For our main experiments, we target all MLP layers and use rank $r=128$.The impact of selecting different layers and modules for fine-tuning is detailed in Appendix \ref{section:Supplementary}. 

\subsubsection{Inference}
During inference, we add the weighted sum of $As$ and $Bs$ to the pretrained weights

\begin{equation}
\text{\textit{inference weights:}}\:\:  \Theta^l_{infer} = \{W^l_{PT}+\sum_i g_i  A^l_iB^l_i\},
\end{equation}

where \(g_i\) are gating weights for each LoRA adapter. 
The gating weights are computed per user query and serve as a global signal shared across all the layers. We compute the gating weights by comparing the user’s query embedding \(f(q)\) with all context keys \(\textbf{K}=[K_1 \: K_2 \: ... \: K_n]\) where $n$ is the total number of memories.

\begin{equation}
\bm{g} = \text{softmax}(\beta f(q)^{\top}\textbf{K}),
\end{equation}

Here $\beta$ is a parameter controlling how spread the gating weights are. We set \(\beta=1\) for the main experiments, and \(\beta=0.1\) for the compositional question experiments.

As mentioned in the previous subsection, we choose $f$ as the average of the token-level internal activation vectors. Since the input must be processed by the base model to generate a response, extracting the embedding from these internal activations incurs no additional computational cost beyond standard inference. Moreover, this design choice ensures that both fine-tuning and retrieval remain fully encapsulated within a single, unified model. 

Given a query $q$ the model generates a sequence of tokens $a_{0:t}$ by iteratively sampling tokens from

\begin{equation}
    p_{\Theta_{infer}}(a_t \mid a_{0:t-1}, q).
\end{equation}

In the main experiments, we use the greedy sampling strategy. Depending on the user query $q$, the model can do different tasks, such as recalling a memory or answering a question related to the memories.

\textbf{Memory Recall: }When testing the model's memorization, we append memory-related questions with an extra prompt ``\textit{Reconstruct the entire story that is related to the above question.}". The model is expected to generate the entire relevant story, rather than answering the question (Figure \ref{Illustration}).

\textbf{QA: } In QA tasks, we found that sometimes the generated answer is too long. To ensure the generation quality and not to confound the question-answering ability with memorization ability, in QA tasks an extra prompt ``\textit{Answer should be no more than one sentence.}" is appended to each question (Figure \ref{Illustration}).

\textbf{Internal RAG (iRAG): } Since the model can recall relevant memories in response to a question, it should also be able to leverage the recalled content to assist in answering the question. We refer to this approach as internal RAG (iRAG). Specifically, given a question, we first append the memory recall prompt to retrieve the full memory associated with the question. The question is then presented to the model again, along with the additional prompt: “\textit{Based on the reconstructed story, answer the following question: \{QUESTION\} Answer should be no more than one sentence.}” iRAG resembles traditional RAG in that it places the relevant knowledge directly into the context window and utilizes in-context learning. However, unlike RAG, the knowledge in iRAG is generated internally by the model rather than retrieved from an external database (see Figure \ref{Illustration}). iRAG can be viewed as a form of chain-of-thought reasoning. Additionally, it resembles the function of memory in traditional associative memory models. For instance, in attractor models long term memories are first being activated by convergence to the fixed point and then uploaded to a working memory module to be further processed according to the demands of the ongoing tasks.  

\subsection{Evaluation}

\subsubsection{Memory Recall}
We evaluate the quality of the recalled memory by computing the cosine similarity between the embeddings (generated by OpenAI text-embedding-3-large) of the model output and the original sample. We refer to it as \textit{recall cos}.

\subsubsection{QA}

In the QA task, we do not require the generated answer to exactly match the correct answer, as the same information can often be expressed in multiple ways (see Appendix \ref{Example-D}). Instead, we use GPT-o3-mini as a judge to determine whether a given response is correct and satisfies the instruction “\textit{Answer should be no more than one sentence.}” (see Appendix \ref{GPT-Judge}). The judge is prompted to produce a binary result. We also report the widely used log probability (log prob) as an additional metric. However, we argue that the GPT judge provides a more reliable evaluation than log prob. Log probability can be misleading: a higher log prob may suggest that the model is more confident, but it does not guarantee that the correct answer is ranked first or even within the top predictions. Moreover, a high log prob can sometimes reflect model collapse rather than genuine correctness in generating the right answers. In contrast, evaluating greedy generation with the GPT judge ensures that both knowledge acquisition (efficiency) and instruction-following ability (generality) are properly assessed.

\subsubsection{General Knowledge Baseline}
To assess whether fine-tuning degrades the model’s general knowledge, we evaluate it on the widely used general knowledge QA dataset, Massive Multitask Language Understanding (MMLU). We report macro accuracy, defined as the average accuracy across four categories: abstract algebra, anatomy, astronomy, and business ethics. Answers are generated in a single step, without chain-of-thought reasoning. The prompt format is detailed in Appendix \ref{section:prompts}.
\subsection{Continual Learning Baselines}

To compare MEGa with non-gated approaches, we include several baselines: continual full-parameter fine-tuning (Full, Full+L2, Full+EWC), continual LoRA fine-tuning (LoRA, LoRA+L2), full-parameter batch learning, and Retrieval-Augmented Generation (RAG).

Continual LoRA entails fine-tuning and merging LoRA adapters into the main weights after each sample \cite{biderman2024lora}. Among these baselines, L2 and EWC (Elastic Weight Consolidation) \cite{kirkpatrick2017overcoming, shan2024order} are common regularization methods for CL. When applying L2 regularization loss, in full-parameter fine-tuning, the L2 loss of the model weights is calculated relative to the model weights before training on the current sample; in LoRA fine-tuning, L2 loss is calculated using the simple LoRA weights. EWC utilizes the second-order error gradient around previous weights. Since, in the continual LoRA setting, the LoRA weights after training on individual samples are merged into the main weights, it is not applicable to use EWC regularization. Thus we only tested EWC in the full-parameter tuning case. We ran a hyperparameter sweep (see Appendix \ref{section:Supplementary}), and chose $\lambda_{L2} = 0.1$ and $\lambda_{\text{EWC}} = 1.0$ in the main experiments.

We also include full-parameter batch learning (mini batch size 2), where the model is trained simultaneously on all available data. Batch learning performance is often considered the ceiling for continual learning methods. As discussed later, the performance gap between batch learning and MEGa provides insights into the potential benefits of leveraging MEGa’s self-recall ability for rehearsal-based learning.

Additionally, we include RAG, which is typically regarded as the ceiling for fine-tuning-based knowledge injection methods. For a fair comparison, we use the same embedding model as MEGa (Llama embeddings). In the main experiments, the top-1 matched sample is put in the context window to generate the answer.

\section{Results}

\subsection{Memory Recall}

\begin{figure}[t]
\begin{center}
\centerline{\includegraphics[width=\columnwidth]{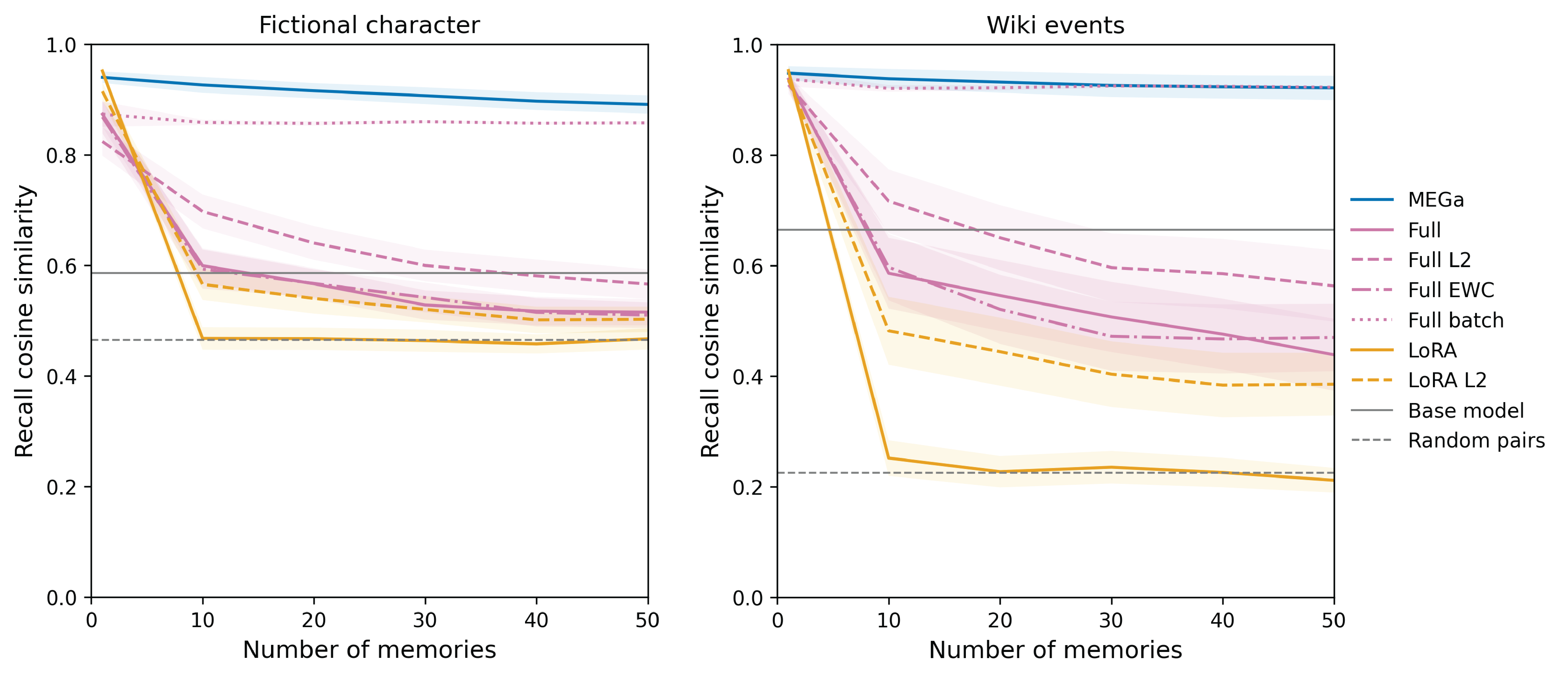}}
\caption{Recall forgetting curve comparing MEGa and other approaches. The model is prompted to recall the fine-tuned sample associated with a given question. Cosine similarity is computed between the recalled text and the original fine-tuning sample. The “Random pairs” baseline represents the average cosine similarity between embeddings of two randomly selected samples (0.465 for the Fictional Character dataset and 0.226 for the Wiki Events dataset).}
\label{forget_recall}
\end{center}
\vskip -0.3 in
\end{figure} 


Table \ref{maintable} presents the evaluation results for MEGa and the baselines. The metric recall cos measures the model’s ability to memorize the fine-tuning samples (see examples in Appendix \ref{appendix:recall}). MEGa excels at recalling relevant memories when cued by questions, achieving a mean cosine similarity of 0.901 on the Fictional Character dataset and 0.921 on the Wikipedia dataset. MEGa selects the correct gate (i.e., the context key with the highest similarity to the question’s source) 85.0\% of the time for the Fictional Character dataset and 87.8\% of the time for the Wiki Events dataset. When the correct gate is chosen, MEGa recalls memories with high fidelity.

While full-parameter batch learning also performs well on this task (0.853 on the Fictional Character dataset and 0.926 on the Wiki Events dataset), all other continual learning (CL) baselines (Full, Full+L2, Full+EWC, LoRA, LoRA+L2) perform substantially worse. These models frequently default to retrieving the most recently seen memory, regardless of the query (see examples in Appendix \ref{appendix:recall}). The base model often fails by refusing to answer, responding with statements like “I don’t have any information about…,” which, although incorrect, sometimes partially overlap with the fine-tuning samples.

The success of MEGa recalling memories also indicates that our fine-tuning procedure retains the model's instruction-following ability, so that the model appropriately processes the retrieval prompt.

The source of the gap between the performance of MEGa and other CL fine-tuning methods is the characteristic of catastrophic forgetting.  We show the forgetting curves computed by the accuracy of retrieving the first sample in the sequence as a function of the length of the sequence in Figure \ref{forget_recall}. All curves exhibit roughly similar good performance initially. However, the CL baselines show severe catastrophic forgetting, although L2 regularization helps mitigate it  to a limited extent (see Appendix \ref{l2curve} for a comprehensive analysis on regularization). In contrast, MEGa shows only mild forgetting. The curve of full-parameter batch learning in Figure \ref{forget_recall} is almost flat. The very small decrease in the ability to retrieve the early stories in response to questions is analogous to shrinking the basins of attraction of memories in attractor networks even below memorization capacity.

\subsection{Question-Answering}

In addition to memorizing the samples, MEGa performs well on direct QA tasks, indicating that the model not only retains the content of the memory but also integrates it into its existing knowledge. MEGa achieves a QA accuracy of 72.53\% on the Fictional Character dataset and 78.03\% on the Wiki dataset, significantly outperforming other continual learning fine-tuning approaches. Its performance is also comparable to that of RAG, which achieves 82.57\% and 88.83\% on the respective datasets—demonstrating that MEGa’s fine-tuning procedure effectively embeds new memories into the model’s weights.

Notably, the base model—prior to any fine-tuning—shows very low QA accuracy (0.13\% on the Fictional Character dataset and 8\% on the Wiki dataset), confirming that the training data introduces novel knowledge (see Table \ref{maintable}).

Figure \ref{forget} shows the forgetting curves for the different fine-tuning methods. Similar to the memory recall task, all methods initially perform well; however, other continual learning (CL) approaches suffer severe degradation in question-answering ability over time. Full-parameter fine-tuning outperforms LoRA without L2 regularization, likely because weight updates are smaller in norms but distributed across more parameters compared to LoRA. Adding L2 regularization improves both full fine-tuning and LoRA fine-tuning, but both still fall significantly short of MEGa’s performance across both datasets. In full-parameter batch learning, QA accuracy exhibits only mild decay as the number of fine-tuning samples increases. MEGa outperforms full-parameter batch learning on the Fictional Character dataset and performs comparably on the Wiki Events dataset, indicating that MEGa’s gating mechanism effectively mitigates forgetting.

Importantly, because MEGa suppresses interference between different memory traces within the fine-tuned model weights, the residual decline in performance with an increasing number of stored memories is primarily due to failures in the softmax-based gating selection mechanism (see Appendix Figure \ref{confusion_curve}). This observation suggests that enhancing the quality of context and query embeddings could further reduce performance degradation. Indeed, we find that employing a superior embedding method substantially reduces forgetting (Appendix Figure \ref{openai_embedding}).

\subsection{Internal RAG}

\begin{figure}[t]
\begin{center}
\centerline{\includegraphics[width=\columnwidth]{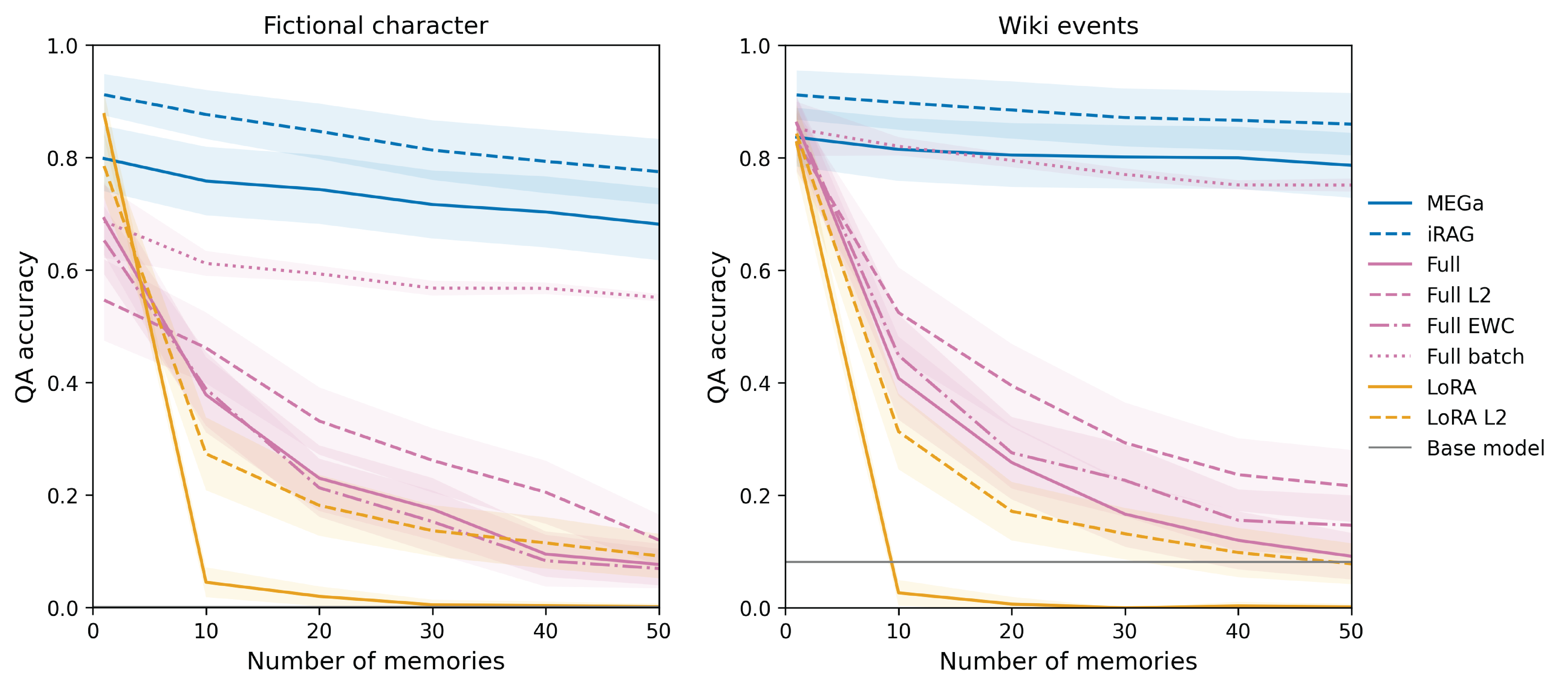}}
\caption{QA accuracy forgetting curve comparing MEGa and other approaches. The curve shows the QA accuracy for questions related to the first sample in the sequence, measured as the model is sequentially trained on additional samples. As more samples are introduced, accuracy on the first sample declines. For the Fictional Character dataset, the base model’s QA accuracy is 0.13\%, overlapping with the x-axis.}
\label{forget}
\end{center}
\vskip -0.3 in
\end{figure}

Inspired by RAG’s high QA accuracy and MEGa’s near-perfect ability to recall relevant samples, we explore iRAG as an alternative way to use MEGa for question-answering. In iRAG, the model first recalls the relevant memory and then answers the question based on the recalled content.  As shown in Table \ref{maintable} and Figure \ref{forget}, iRAG further boosts MEGa’s QA performance, achieving 80.67\% accuracy on the Fictional Character dataset and 84.70\% on the Wiki Events dataset—effectively closing the performance gap with RAG.

\subsection{General Knowledge Retention}
\begin{figure}[H]
\begin{center}
\centerline{\includegraphics[width=\columnwidth]{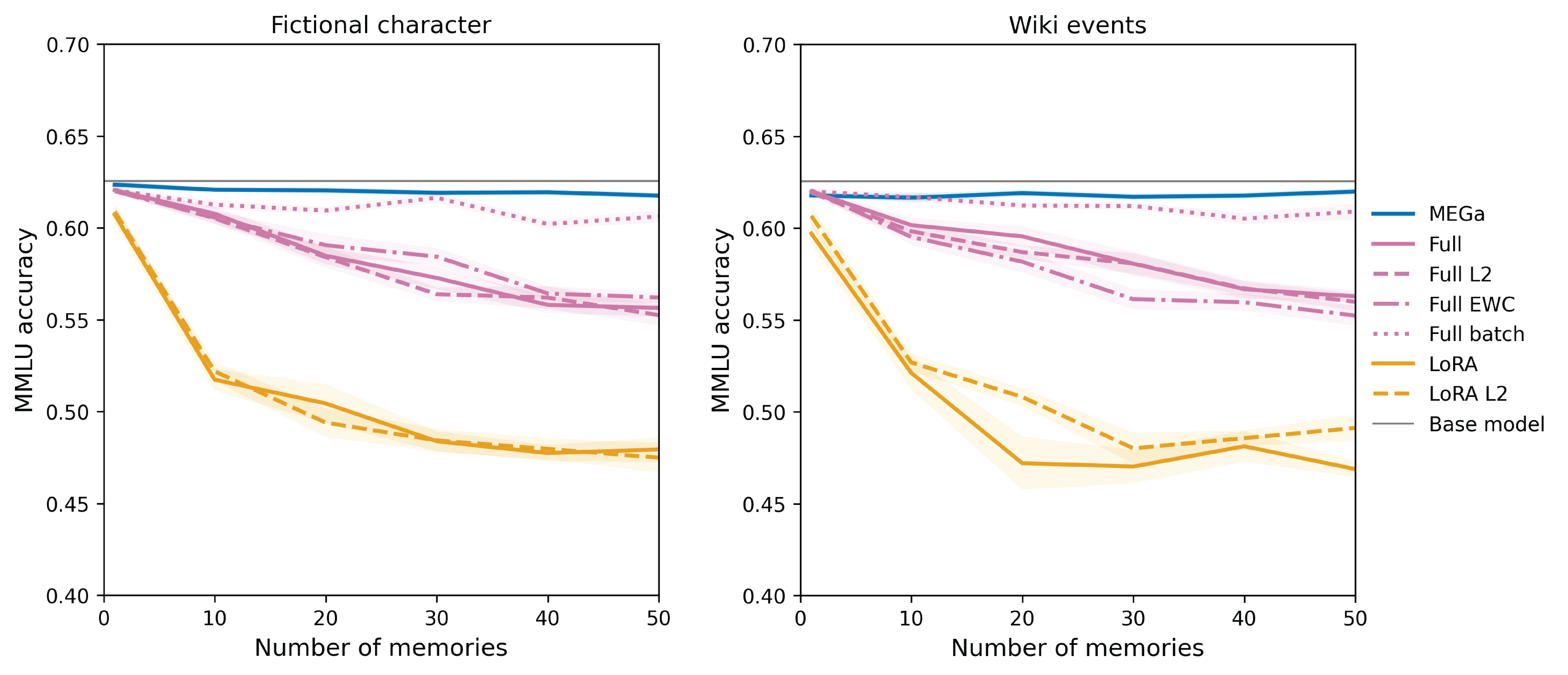}}
\caption{MMLU accuracy during continual fine-tuning.}
\label{MMLUcurve}
\end{center}
\end{figure}

We also evaluate the potential degradation of the model’s general language capabilities resulting from fine-tuning. After training on both datasets, MEGa maintains MMLU accuracy (61.75\% and 61.99\%), closely matching the performance of the base model, whereas all other continual learning (CL) fine-tuning methods exhibit a noticeable decline (see Table \ref{maintable} and Figure \ref{MMLUcurve}). These results suggest that MEGa effectively integrates new information while preserving prior general knowledge and instruction-following abilities—key features for robust continual learning systems.

\subsection{Composite Knowledge}

MEGa can mix different LoRA adapters by applying softmax-weighted gating, enabling the combination of knowledge from multiple memories. To test this capability, we evaluate whether such a mixture can correctly answer questions that require information from two separate samples.

We find that full-parameter batch learning performs best on compositional questions, with MEGa performing comparably. Among continual learning methods, MEGa significantly outperforms all other CL baselines (see Table \ref{table:composition}). Example compositional questions and model answers are provided in Appendix \ref{Appendix-E.1}.

While model merging has been widely used to combine task-specific models for computational efficiency \cite{yang2024model, wang2024wise}, our results provide the first demonstration that knowledge embedded in separate LoRA modules can be successfully integrated to answer compositional questions through simple weight merging.

\begin{table}[!ht]

\vskip 0.1in
\begin{center}
\scriptsize
\begin{sc}
\begin{tabular}{l|c|c}
\toprule
& \multicolumn{1}{c|}{Fictional} & \multicolumn{1}{c}{Wiki} \\
Methods & Character & dataset \\
\midrule
MEGa         & 49.6\% & 70.4\% \\
Full         & 9.6\%  & 20.0\% \\
Full L2      & 26.0\% & 35.6\% \\
Full EWC     & 8.4\%  & 16.4\% \\
Full Batch   & \textbf{54.4\%} & \textbf{75.2\%} \\
LoRA         & 0.4\%  & 0.4\%  \\
LoRA + L2    & 8.4\%  & 18.4\% \\
\bottomrule
\end{tabular}
\end{sc}
\caption{Compositional QA Accuracy.}
\label{table:composition}
\end{center}
\vskip -0.1in

\end{table}





\section{Discussion}

In this study, we present a framework, MEGa, for sequentially injecting new pieces of knowledge into a pretrained LLM. 
Our goal is twofold: address current challenges in continual knowledge injections in LLMs and AI agents, and develop LLM-based viable models of human memory functions. Some of the challenges met in this work arise from the features of current LLMs such as sensitivity to specific prompts and the varying quality of embeddings. Others are more general and have implications for any intelligent systems, such as mitigating the risk of catastrophic forgetting (CF). Continual learning remains a challenge for ML systems, in contrast to the impressive abilities of humans as lifelong learners, accumulating new memories and knowledge through sequential experiences with their environment. MEGa injects individual memories into  fine-tuned LoRA modules and mitigates CF through an integrated gating mechanism. MEGa achieves performance comparable to, or surpassing,  RAG and full-parameter batch learning, which are considered the ceiling performance for knowledge editing and continual learning, respectively. Notably, MEGa stores memories within network weights rather than relying on an external database like RAG, thus providing a more biologically plausible model of human memory. 

We have constructed and studied two datasets.  The Wiki events dataset is widely adopted in LLM knowledge injection studies \cite{ovadia2023fine, mecklenburg2024injecting, zhang2024comprehensive, mecklenburg2024injecting, thede2025understanding}. These types of memories resemble factual knowledge, a form of human semantic memory. In contrast, the autobiographic nature of the stories in the fictional character dataset makes them candidates for the study of human episodic memories and also crucial for AI personalization, alignment, and agency. Despite their importance, datasets resembling everyday life memories about a person were rarely used in LLM knowledge injection studies.

Not surprisingly, the two datasets have different statistical structures. The fictional character stories are more correlated than the wiki events, hence suffer more from interference in the gating selection operation (Appendix Figure \ref{embedding_layers}), even with strong embedding models (Appendix Figure \ref{confusion_curve}, \ref{openai_embedding}). Another interesting finding is that  fine-tuning later MLP layers is more effective to inject fictional character knowledge; while fine-tuning early MLP layers is more effective for the Wiki event knowledge, suggesting that knowledge from the two datasets, due to their distinct nature, might be located in different layers. This hypothesis needs to be validated by further analysis including ablation experiments. 



Our current feasibility study uses relatively small scale memory data. As indicated by our results (Appendix Figure \ref{confusion_curve} \ref{openai_embedding}), scaling up the model is possible if one uses an LLM with better embeddings than the present one. In addition,  both in humans and AI agents, episodic memories enter the neural systems as sensory experiences. Our preesent 'autobiographic' stories should be interpreted as internal summaries of the experiences. Extending our model to incorporate multimodal episodic memories in MEGa is an important future research goal. 

One limitation of MEGa is that its parameter count grows linearly with the number of training samples, as each new memory requires additional LoRA adapter. This scalability issue can lead to increased computational and storage demands. This might be partially resolved by using shared LORA weights and post learning pruning methods. 

A promising future direction is to gradually distill LoRA weights into the base model weights. This entails a rehearsal process in which gating units are activated repeatedly, generating a spontaneous reconstruction of one or few stored memories at a time, and then triggering a slow fine-tuning of the base model. Our batch baseline finetuning, which uses fine-tuning steps of mini-batches of 2 memories, can be thought of as implementing such a rehearsal process. To complete this transfer, it is important to complement this rehearsal process with a gradual phasing out of some LoRA adapters depending on age, frequency of use, or other saliency criteria. 
Incorporating rehearsal-based memory transfer will make the model similar to the 
complementary memory systems hypothesis for human long-term memory. Here, the gating systems and associated LoRA weights correspond to the fast learner (``hippocampus") while the rehearsal-triggered fine-tuning of the base weights correspond to the slow learner (``cortex") \cite{mcclelland2020integration}. 
The gating operation in MEGa is also reminiscent of the indexing theory of hippocampal memory \cite{teyler1986hippocampal}.

In the present version of MEGa, individual stories are generated in advance as separate independent stories, whereas both computational considerations and evidence from studies of human memory, indicate that chunking should be a dynamic flexible process which can reduce redundancy or shared context. A simple improvement would be to add an option of merging and splitting of stories depending on the similarity between them. In particular, a new memory might update a similar existing memory rather than being stored as a distinct event. Furthermore, it is promising to organize memory modules as nodes in a graph, similar to approaches in RAG \cite{gutierrez2024hipporag, han2024retrieval}, where gating can be guided by traversing a knowledge graph structure. This would allow more efficient memory reuse, compositional reasoning, and structured retrieval based on relationships between memories.

\section*{Acknowledgments}

We acknowledge the support of the Swartz Foundation, the Kempner Institute for the Study of Natural and Artificial Intelligence at Harvard University, the Office of Naval Research (ONR) grant No.N0014-23-1-2051, and the Gatsby Charitable Foundation. We have benefited from helpful discussions with Jorin Overwiening, Qianyi Li, Jingxuan Fan, Anya Ben Yakov, and Isaiah Kletenik.

\bibliography{reference}
\bibliographystyle{icml2025}

\newpage
\appendix
\onecolumn
\section{Code availability}
MEGa code and datasets are in the repository \url{https://github.com/xup5/MEGA}.

\section{Supervised Fine-Tuning}
\label{section:superpervised_finetuning}

\begin{figure}[H]
\begin{center}
\centerline{\includegraphics[width=0.7\columnwidth]{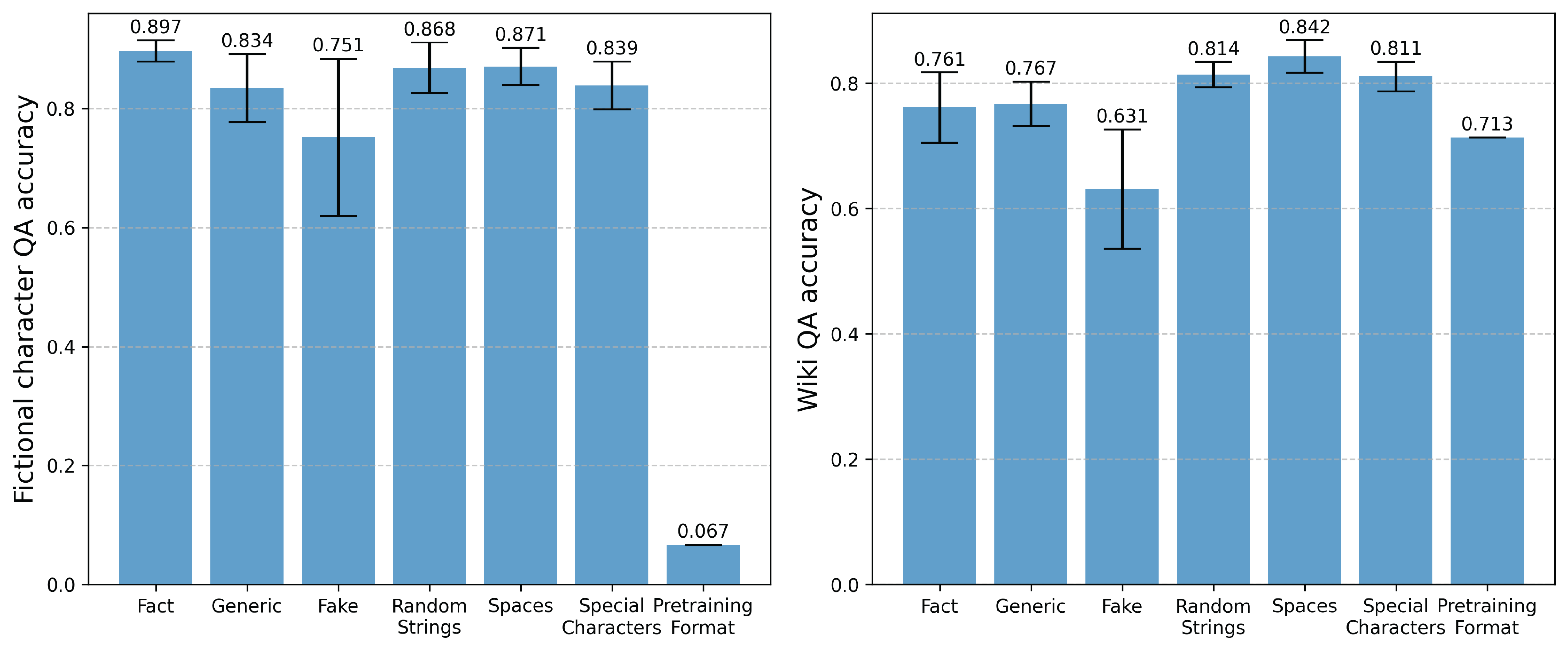}}
\caption{We found that the effectiveness of injecting new knowledge into instruct model depends on the ``fine-tuning prompt". The category ``fact" represents the prompts (n=15) that indicate the finetuning sample is a fact, for example ``Please tell me a story that you memorized:"; the category ``generic" represents the prompts (n=7) that do not imply the factuality of the sample, for example ``Generate:"; the category ``fake" represents the prompts (n=10) that indicate the sample is fictional, for example ``Please make up a fake story:". We also have categories ``random strings" (n=24), ``spaces" (n=4), and ``special characters" (n=10). The accuracy is the GPT judge's results on 750 questions related to 250 samples in each dataset. Fine-tuning is done per sample (no continual learning). The error bar represents the standard deviation among prompts.}
\label{prompts}
\end{center}
\end{figure}

A priori, it is not obvious what format to use when fine-tuning instruct models on raw knowledge paragraphs. We first tried to fine-tune with the pretraining instruct format (Supplementary Figure \ref{prompts}). We found that such a way induces some question-answering ability on the Wiki event dataset, but fails on the fictional character dataset, as the model almost always refuses to answer. 

As such, like in the supervised fine-tuning, we propose to construct QA pairs where the answers are the raw knowledge paragraphs. We constructed 70 different questions/prompts to pair with the raw knowledge paragraphs. We refer to them ``fine-tuning prompts." They are categorized into 6 categories: ``fact", ``generic", ``fake", ``random strings", ``spaces", and ``special characters".

On both datasets, ``fact" prompts perform significantly better than ``fake" prompts regarding QA accuracy. One possible explanation is that if the fine-tuning text is presented in a place where the model's knowledge is supposed to be, then it is easier for the model to update some targeted weights that are related to storing this knowledge.

We choose to use ``Please tell me a story that you
memorized:" as the fine-tuning prompt in all main experiments.

\begin{verbatim}
Fact based prompts:
"Reconstruct the story:"
"Please reconstruct the story:"
"Tell me a story that you memorized:"
"Please tell me a story that you memorized:"
"Recall a memorized fact:"
"Retrieve the known information:"
"Output what you remember clearly:"
"Repeat the exact fact you know:"
"Please recall stored information:"
"Tell me what you know:"
"Explain the known detail:"
"Provide the stored information:"
"Share a known fact:"
"Produce information already known:"
"Give me a fact known to you:"

Generic prompts:
"Tell me something:"
"Output something:"
"Go ahead:"
"Generate:"
"Please proceed:"
"Say something meaningful:"
"Response:"

Fake based prompts:
"Make up a fake story:"
"Please make up a fake story:"
"Tell me a fake fact:"
"Make up a completely new fact:"
"Invent an imaginary fact:"
"Create a completely original fact:"
"Generate something entirely fictional:"
"Tell a fact that’s not true:"
"Fabricate information on the spot:"
"Share something completely unfamiliar:"

Random string:
"U9iG2"
"RYD0N"
"ySlui"
"sRPI5"
"jI79X"
"r7qlZl51El"
"2xYq18fV0U"
"LKGlHIeLlw"
"B834aabqWT"
"oVAj1weRvA"
"m8TpLXMZR1dQyU7"
"UX9bmF6Pppwd25m"
"DruwGzgHGVLqUqu"
"jvcObYlUPSH3Yr2"
"avPtYk9eZAHDPgv"
"F1hXxX2JsHt3zck4RVK4"
"FMXCVUY9Mo0gBGk0UCCd"
"kW7OiLueZZyY3Qi1Ss3m"
"0T9JHWpwki1Zxz45Wu3N"
"DiY7FEbnaYTnQgMjYDik"
"11111111111111111111"
"22222222222222222222"
"####################"
"!!!!!!!!!!!!!!!!!!!!"

Spaces:
" "
"     "
"          "
"                    "

Special characters:
"(*&^%$#@!)"
"+_)(*&^%$#@!~"
"/.,;[]=-0987"
"/{/}<>:"
"|\\//||\\//||"
">?<}{|+=_-)(*&^%$#@!~"
"!@#%^&*()_+/{/}|:?><,./;'[]\\-="
"/}/{[]:;'.,<>/?|\\~!@#$%^&*()-_=+`"
"$#@!%^&*()_+=-[]/{/}|:;<>,.?/~`:"
"/}/{|:><?/.,';[]=-0987654321"
\end{verbatim}



\section{Supplementary Tables and Figures}
\label{section:Supplementary}





\begin{table}[h]
\label{notmaintable}
\begin{center}
\begin{small}
\begin{sc}
\begin{tabular}{p{0.1\linewidth} | >{\scriptsize}p{0.11\linewidth} >{\scriptsize}p{0.11\linewidth} >{\scriptsize}p{0.11\linewidth} | >{\scriptsize}p{0.11\linewidth} >{\scriptsize}p{0.11\linewidth} >{\scriptsize}p{0.11\linewidth}}
\toprule
& \multicolumn{3}{c|}{Fictional character} & \multicolumn{3}{c}{Wiki events} \\
& QA Acc \% & Log prob & Recall cos & QA Acc \% & Log prob & Recall cos \\
\midrule
All MLP  & \textbf{89.60\% ± 1.16\%} & \textbf{-1.5219 ± 0.1185} & 0.9560 ± 0.0063 & \textbf{81.07\% ± 2.17\%} & \textbf{-1.5146 ± 0.1263} & 0.9564 ± 0.0033\\
All Attn & 86.53\% ± 5.21\% & -1.9229 ± 0.1160 & 0.9409 ± 0.0060 & 72.80\% ± 2.44\% & -1.8268 ± 0.1549 & 0.9568 ± 0.0085\\
All Lyrs & 81.07\% ± 1.72\% & -1.7159 ± 0.0803 & \textbf{0.9621 ± 0.0041} & 51.60\% ± 3.62\% & -1.8253 ± 0.1608 & \textbf{0.9584 ± 0.0037}\\
Early MLP& 51.33\% ± 2.89\% & -2.2454 ± 0.1109 & 0.8108 ± 0.0347 & 73.17\% ± 6.87\% & -1.7015 ± 0.0923 & 0.9304 ± 0.0049\\
Mid MLP  & 56.13\% ± 3.80\% & -2.0896 ± 0.0476 & 0.9034 ± 0.0191 & 76.80\% ± 4.01\% & -1.6900 ± 0.0602 & 0.9254 ± 0.0149\\
Late MLP & 58.53\% ± 4.68\% & -2.4736 ± 0.0705 & 0.8564 ± 0.0100 & 60.13\% ± 2.71\% & -1.9014 ± 0.1091 & 0.8583 ± 0.0092\\
\bottomrule
\end{tabular}
\end{sc}
\end{small}
\end{center}
\caption{Ablation analysis on fine-tuning layers with LoRA. Fine-tuning and evaluation are done per sample (no continual learning). Early MLP are layers 1-10; Middle MLP are layers 10-21; Late MLP are layers 21-32. Mean and standard deviation across 20 dataset partitions are reported.}
\end{table}


\begin{table}[h]
\label{l2curve}
\begin{center}
\begin{small}
\begin{sc}
\begin{tabular}{p{0.15\linewidth} | >{\scriptsize}p{0.11\linewidth} >{\scriptsize}p{0.11\linewidth} >{\scriptsize}p{0.11\linewidth} | >{\scriptsize}p{0.11\linewidth} >{\scriptsize}p{0.11\linewidth} >{\scriptsize}p{0.11\linewidth}}
\toprule
& \multicolumn{3}{c|}{Fictional character} & \multicolumn{3}{c}{Wiki events} \\
& QA Acc \% & Log prob & Recall cos & QA Acc \% & Log prob & Recall cos \\
\midrule
LoRA l2 = 0.001  & 1.03\%±1.45\% & -3.949±0.435 & 0.476±0.042 & 2.30\%±2.82\% & -2.585±0.435 & 0.242±0.058 \\
LoRA l2 = 0.01  & 6.40\%±4.79\% & -3.444±0.490 & 0.495±0.046 & 7.07\%±6.59\% & -2.406±0.439 & 0.295±0.085 \\
LoRA l2 = 0.1  & 17.03\%±7.01\% & -3.324±0.409 & 0.525±0.049 & 19.63\%±9.59\% & -2.269±0.353 & 0.447±0.144 \\
\textbf{LoRA l2 = 1}  & 20.07\%±3.39\% & -3.540±0.417 & 0.769±0.022 & 29.50\%±5.70\% & -2.740±0.317 & 0.818±0.038 \\
LoRA l2 = 10  & 9.12\%±3.11\% & -4.480±0.476 & 0.724±0.033 & 13.90\%±4.92\% & -3.374±0.378 & 0.771±0.051 \\
Full l2 = 0.001  & 17.37\%±8.20\% & -3.001±0.318 & 0.539±0.061 & 21.50\%±11.57\% & -2.008±0.331 & 0.510±0.179 \\
Full l2 = 0.01  & 18.20\%±8.90\% & -3.051±0.367 & 0.543±0.062 & 22.10\%±13.09\% & -2.040±0.270 & 0.525±0.174 \\
Full l2 = 0.1  & 27.73\%±10.07\% & -3.198±0.325 & 0.581±0.071 & 33.20\%±9.51\% & -2.117±0.309 & 0.602±0.156 \\
\textbf{Full l2 = 1}  & 29.50\%±3.76\% & -3.677±0.269 & 0.783±0.024 & 45.43\%±5.58\% & -2.444±0.298 & 0.821±0.044 \\
Full l2 = 10  & 11.07\%±4.01\% & -4.337±0.340 & 0.759±0.031 & 33.77\%±3.86\% & -2.822±0.310 & 0.829±0.022 \\
Full EWC = 0.01 & 13.33\%±7.66\% & -3.069±0.376 & 0.558±0.064    & 19.23\%±12.97\% & -1.969±0.371 & 0.545±0.162\\
Full EWC = 0.1 &  12.83\%±9.02\% & -3.012±0.291 & 0.542±0.063    & 19.23\%±10.53\% & -1.994±0.355 & 0.609±0.140\\
\textbf{Full EWC = 1} &  14.07\%±7.76\% & -2.818±0.286 & 0.542±0.064    & 21.63\%±13.94\% & -2.023±0.320 & 0.544±0.149\\
Full EWC = 10 & 13.57\%±8.49\% & -2.918±0.368 & 0.535±0.068    & 18.30\%±9.76\%  & -2.088±0.395 & 0.566±0.144\\
Full EWC = 100 & 13.63\%±9.36\% & -2.953±0.399 & 0.535±0.050    & 18.43\%±13.40\% & -2.032±0.295 & 0.533±0.146\\
\bottomrule
\end{tabular}
\end{sc}
\end{small}
\end{center}
\caption{Hyperparameter search for regularization coefficients. The evaluation is done at the end of sequentially training on 50 samples. Mean and standard deviation across 20 dataset partitions are reported. Though L2 coefficient $1.0$ yields the best continual learning performance, we choose to use $0.1$ in the main text, because when L2 coefficient $1.0$ failed to inject a single fictional character stories (almost always refuses to answer). A possible explanation is that the pretrained instruct model is more restricted to answer questions related personal information due to RLHF safety guideline. Knowledge injection by finetuning needs to perturb the model sufficiently large to break its refusal behavior. We show a comparison between L2 coefficient $0.1$ and $1.0$ in the Figure \ref{forgetting_curve_l2}. We choose to use EWC coefficients of $1.0$ in the main experiments.}
\end{table}

\begin{figure}[H]
\begin{center}
\centerline{\includegraphics[width=0.55\textwidth]{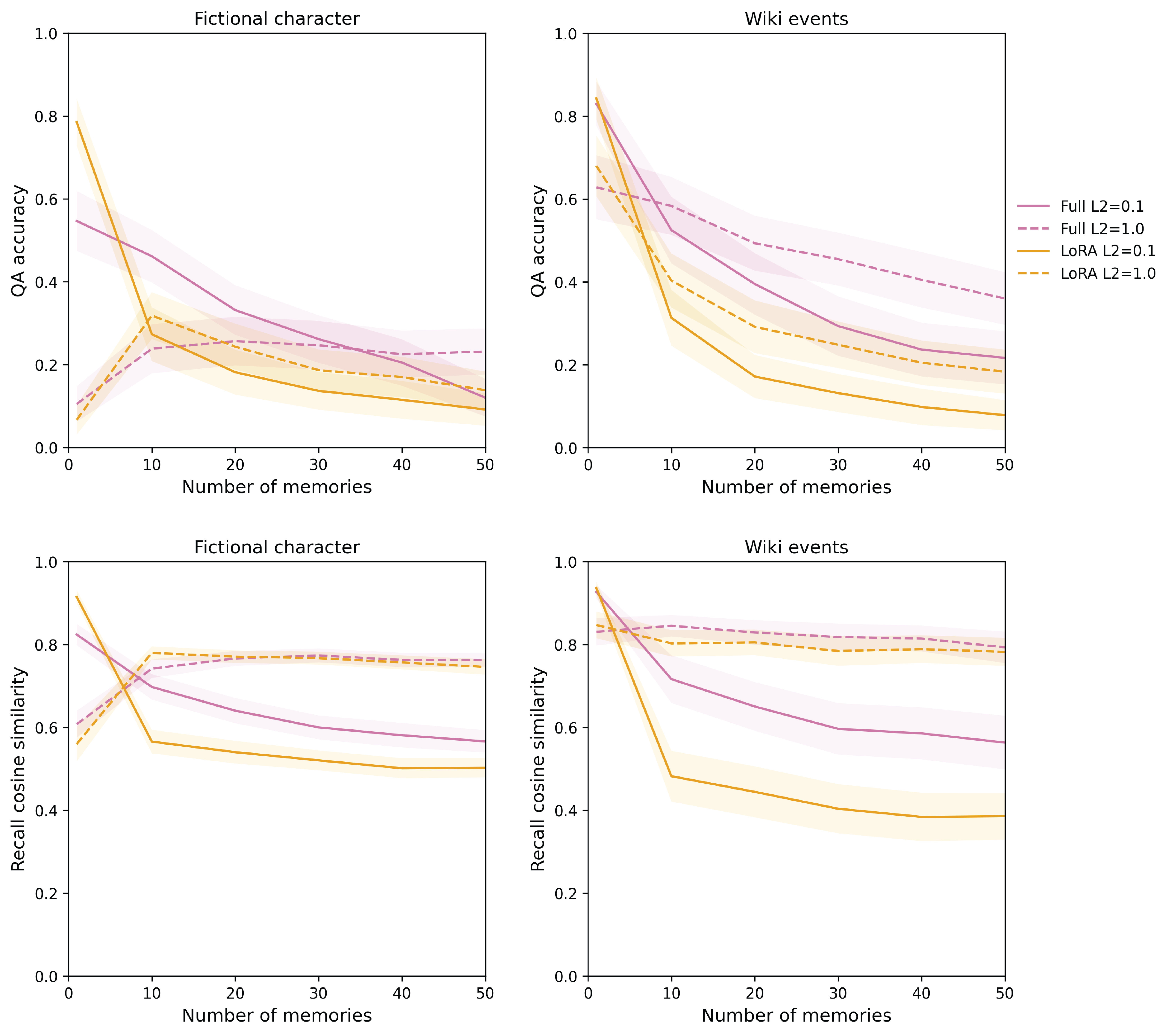}}
\caption{Comparison between L2 coefficient $0.1$ and $1.0$. L2 coefficient $1.0$ elicits an interesting increasing trend in the QA accuracy forgetting curve. The model refuses to answer the question after finetuning on the first sample. The refusal behavior is removed after finetuning on more samples. One possible explanation is the model needs sufficient large perturbation to break its safety guidance gained from RLHF.}
\label{forgetting_curve_l2}
\end{center}
\end{figure}

\begin{figure}[H]
\begin{center}
\centerline{\includegraphics[width=0.55\textwidth]{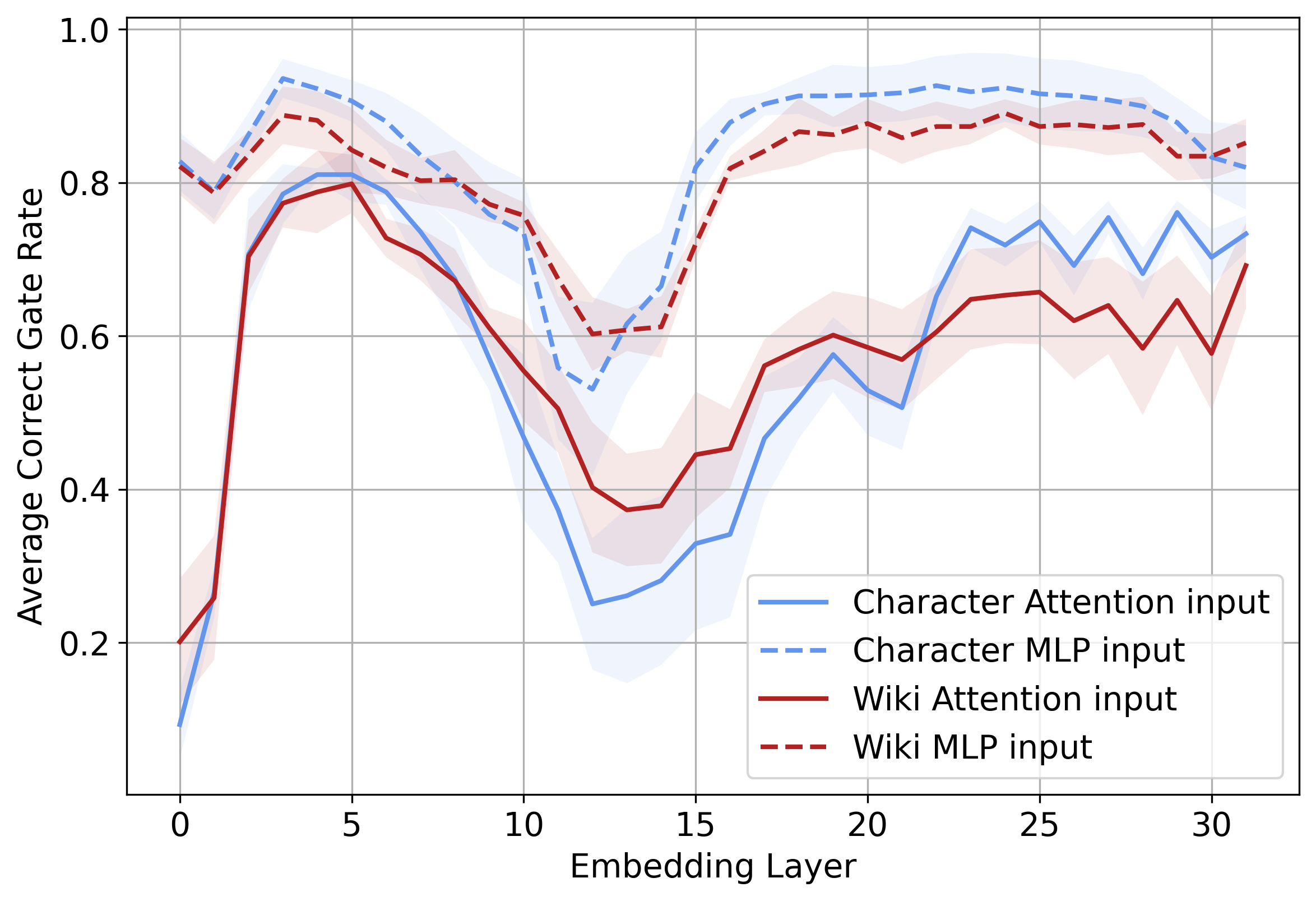}}
\caption{Analysis on the embedding quality in Llama intermediate layers. Correct gate rate is computed as the rate of question embeddings which have the maximum inner product with the relevant passage embedding. It is computed for each dataset partition, which has 50 passage samples. Shaded area is the standard deviation across 20 dataset partitions. We found that embeddings at the input of MLP layers are better than at the input of attention layers. We choose to use last MLP layer input as embedding in the main experiments of MEGa.}
\label{embedding_layers}
\end{center}
\end{figure}

\begin{figure}[H]
\begin{center}
\centerline{\includegraphics[width=0.7\textwidth]{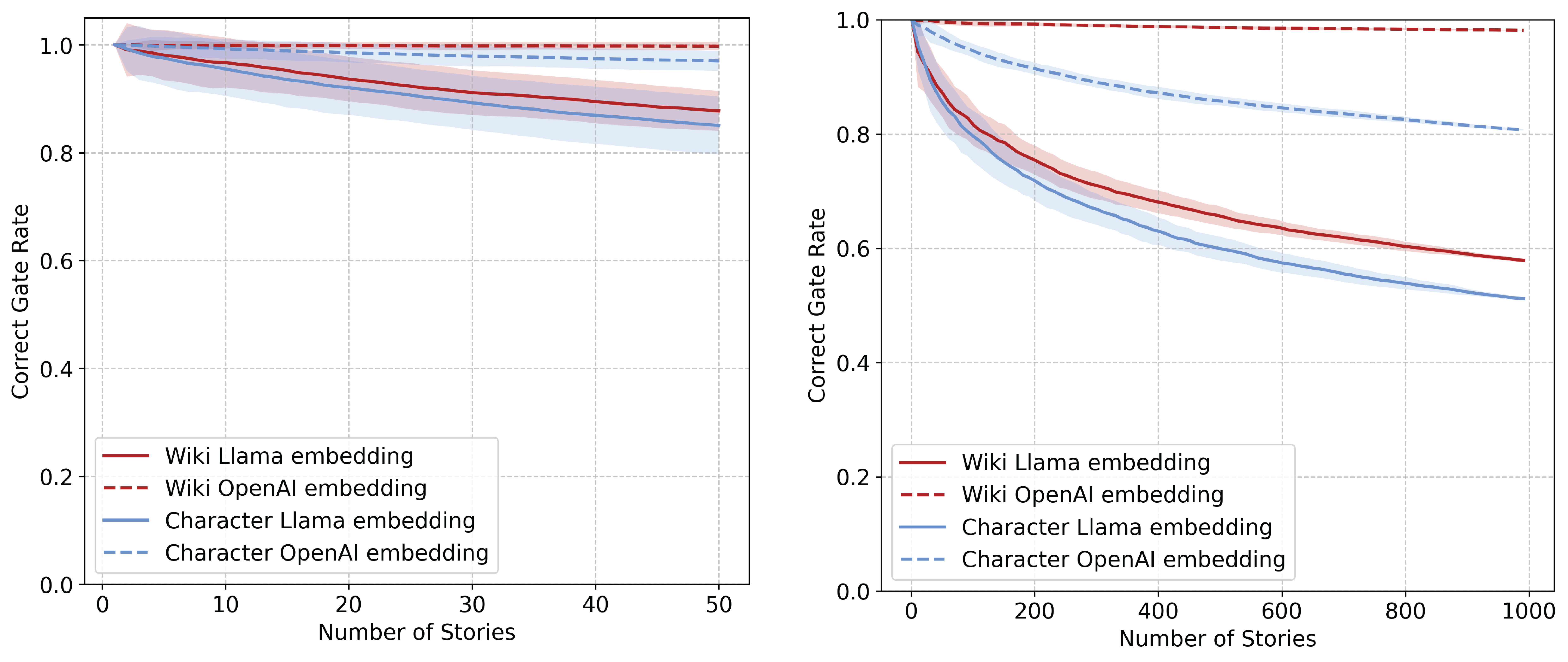}}
\caption{Due to the modular design, MEGa's performance can be instantly boosted by replacing Llama embedding by other embedding models. We compare the performance of MEGa with Llama embeddings and OpenAI text-embedding-3-large embeddings.}
\label{confusion_curve}
\end{center}
\end{figure}

\begin{figure}[H]
\begin{center}
\centerline{\includegraphics[width=0.7\textwidth]{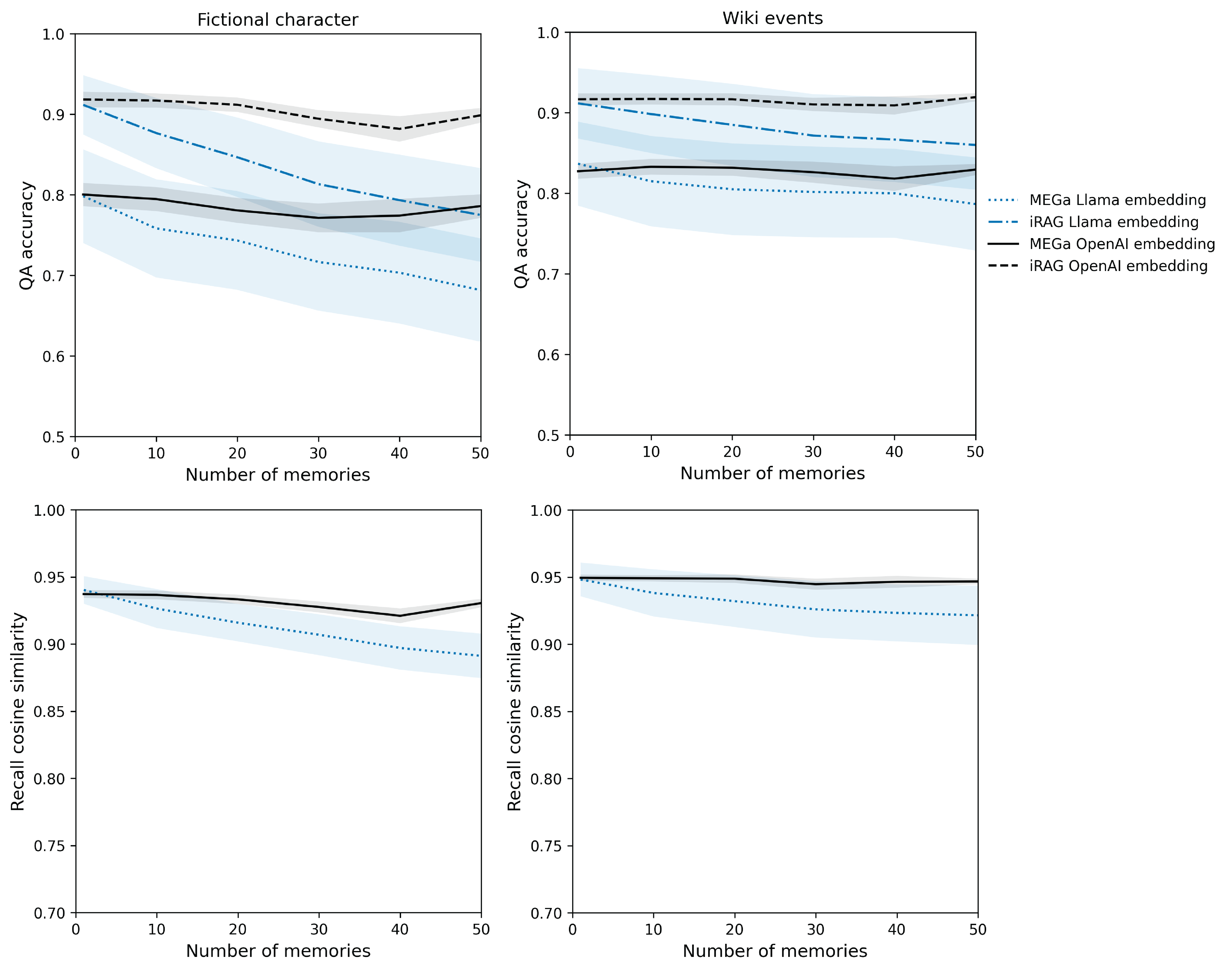}}
\caption{Comparing Llama embedding with OpenAI text-embedding-3-large embedding. The datasets used in the left plot are 20 dataset partitions, each with 50 samples (i.e. stories), as in other experiments. In the right plot, to test how does the correct gate rate scaling with number of stories beyond 50, we concatenate all dataset partitions into one. For the fictional character dataset, we change the character name to a same name before concatenating the partitions.}
\label{openai_embedding}
\end{center}
\end{figure}



\section{Fine-Tuning Settings} 
\label{section:Fine-Tuning Settings}
All main experiments were done using 2×H100-80GB GPUs. 
As per the MLP versus attention layer ablations, we fine-tune only the MLP modules in each of 
Llama-3.1-8B-Instruct's 32 layers i.e., $\{W^{(l)}_{\text{up-proj}}, W^{(l)}_{\text{down-proj}}\}_{l = 1}^{32}$. Across all experiments involving LoRA, we fine-tune each LoRA adapter for 10 epochs with a learning rate $\eta = 3 \cdot 10^{-5}$ using the AdamW optimizer. For each LoRa adapter, as per the rank ablations in Appendix \ref{section:Supplementary}, we take $r = 128$. Following \citet{kalajdzievski2023rank}, who theoretically show that setting the the LoRA scaling factor $\gamma \in \Theta(\frac{1}{\sqrt{r}})$ results in a rank-stabilized LoRA, for all experiments we take the LoRA scaling factor $\gamma = \frac{\alpha}{\sqrt{r}} = \sqrt{128}$ by setting the hyperparameter $\alpha = r = 128$.

For the full fine-tuning baseline, we keep all applicable hyperparameters the same as above besides the learning rate, which we opt for $\eta = 1 \cdot 10^{-5}$, a smaller learning rate than the LoRA setting since the norm of the original weights is larger than LoRA initialization weights.

In all experiments, for the inference sampling strategy, we set $\text{do\_sample} = \text{False}$ i.e., we use greedy sampling.

\section{Prompts}
\label{section:prompts}
\subsection{Dataset Generation}

\subsubsection{Fictional Character Dataset Stories}
To generate 20 synthetic datasets, we use the \texttt{gpt-4.5-preview-2025-02-27} model with a temperature setting of 1.0. Each dataset consists of 50 short stories centered around events in the life of a single fictional character. To minimize potential overlap with the model’s pretraining data, the prompt explicitly includes the instruction to ``include details unlikely to be public knowledge," encouraging generation of novel and highly specific content. We also provide gpt-4.5 with 5 handwritten examples. Names for the characters are generated using a random name generator, and we have 10 male and 10 female characters across the datasets. For each character, the model is prompted in batches of 10 stories at a time, iterated to produce the full set of 50 stories.
\begin{lstlisting}
    prompt = f"""
Based on the five examples below, create 10 more stories about the fictional character {firstname} {lastname}. 

Each story must:
- Consist of exactly 2 short sentences.
- Be packed with unique personal details, including specific names, dates, locations, and vivid scenarios.
- Cover distinctly different topics to ensure diversity.
- Include details unlikely to be public knowledge, maintaining a personal and authentic feel.

Examples:
1. At age 13, {firstname} {lastname} celebrated his bar mitzvah at Jerusalem's Western Wall, reciting the Torah passage Nitzavim. Afterwards, {firstname} joyfully danced the horah with his uncles.

2. When {firstname} {lastname} was 17, his precise three-point shooting earned him the title of MVP after securing the high school basketball championship in Edison, New Jersey. His game-winning shot in the final seconds made him a school hero overnight.

3. {firstname} {lastname}, at 15, solved the final geometry problem at the 2019 regional math competition with a clever proof that astonished even the judges. His solution earned him a coveted spot in the national round later that year.

4. {firstname} {lastname} visited Switzerland with college friends, attempting skiing for the first time in the Alps. On his third day, overestimating his skills, he tackled an intermediate slope, fell awkwardly, and fractured his wrist.

5. During a soccer match in Wayne, New Jersey, in October 2012, {firstname} {lastname} impressively scored five goals by halftime, single-handedly dominating for the Tigers. To rebalance the game, amused referees switched him onto the opposing team, the Lions.
"""
\end{lstlisting}

\subsubsection{Generating Paraphrases}
To generate 9 paraphrases for the training dataset, we prompt \texttt{gpt-4.5-preview-2025-02-27} as follows:
\begin{lstlisting}
prompt = f"""Your task is to paraphrase a text paragraph. The paragraph is given below. 
Make sure to keep the same meaning but change the wording. Do not change any factual 
information. Try to keep roughly the same length as the original text. Provide exactly 
9 different paraphrases for the given text by numbering them 'Paraphrase 1:', 
'Paraphrase 2:', etc.

Input paragraph:
{passage}
"""
\end{lstlisting}

\subsubsection{Generating QA }
\label{QA-Generation}
To generate 3 QA pairs per memory for evaluation for both datasets, we prompt \texttt{gpt-4.5-preview-2025-02-27} as follows:
\begin{lstlisting}
f"""Your task is to generate 3 question and answer pairs based on a given passage below. Make sure to provide AMPLE context in the question, including information from the original passage as context. Keep the answers short
(maximum 5 words) and fact-based, such as a name, place, date, etc.. Return a JSON formatted string with one key, called qa-data, and a list of (question, answer) tuples.
Input paragraph:{story}
"""
\end{lstlisting}

\subsection{Compositional Question Dataset Construction}
To construct compositional questions, we prompt \texttt{o3-mini-2025-01-31} as follows. We split each partition (50 stories) into five batches, each consisting of 10 stories. For each batch, we generate 5 questions, as per the prompt below. Thus, we ultimately have 25 questions per partition; we opt for 10 fictional character partitions and 10 Wiki partitions for a total of 500 compositional questions.
\begin{lstlisting}
f""" You are given a list of 10 passages about a person named {firstname} {lastname}. Each passage describes an event in his life. Here they are:
{ten_stories}

Your task: 
- Generate exactly 5 questions that each require information from exactly two distinct passages above.
- Each question should focus on short factual details such as name, location, date, or age (so the answer is typically 1-3 words) stemming from either similarities or differences between these factors, with an emphasis focusing on similarities.
- Make sure to reference the relevant details from precisely two passages in each question and provide ample context from the content of the passages, without mentioning the passage numbers. Make sure the answer cannot be gleaned from the question only -- it must be obtained via knowledge of the passages. 

**Output format**:  
Return your final output strictly as a JSON-like list of tuples:  
[
  ["QUESTION_1", "ANSWER_1"],
  ["QUESTION_2", "ANSWER_2"],
  ["QUESTION_3", "ANSWER_3"],
  ["QUESTION_4", "ANSWER_4"],
  ["QUESTION_5", "ANSWER_5"]
]

**Example**:
For instance, if two passages are:
1) 'At age 9, Kelly Dash won his elementary school's annual talent show in Montclair, New Jersey, juggling five tennis balls to surprised applause. His proud grandmother Charlotte captured the entire performance on her cherished handheld camcorder.'
2) 'At age 11, Kelly Dash nervously performed an original poem titled "Fireflies in June" at the Passaic County youth poetry slam held in Paterson, New Jersey. His heartfelt delivery earned him third place, plus a congratulatory handshake from local hero poet Elena Martinez.'

then a valid output would be
[
  ["Did Kelly Dash win his elementary school's annual talent show at age 9 in the same state he nervously performed an original poem titled 'Fireflies in June'?", "Yes, New Jersey"]
]

Now produce 5 such (question, answer) pairs, referencing the passages by looking at exactly two of them each time.
"""
\end{lstlisting}
\label{compquestions}

\subsection{GPT judge} 
\label{GPT-Judge}
To access the correctness of the fine-tuned model's responses to questions about a given story, we employ a GPT-based judge (\texttt{o3-mini-2025-01-31}) with a temperature setting of 1.0, following the prompt similar to \cite{mecklenburg2024injecting}.

\begin{lstlisting}
prompt = (
f"You are evaluating a prospective answer to a question on a given article. "
f"Your grading is binary: give 1 if and only if the prospective answer is correct (that is, the prospective answer contains the actual correct answer) and the prospective answer is no more than 1 sentence long; give 0 if any of these two conditions are not satisfied."
f"Your output is either 0 or 1, no other information should be in the output.\n\n"
f"The article: {story}\n\n"
f"The question: {question}\n\n"
f"The correct answer: {actual_ans}\n\n"
f"The prospective answer: {model_ans}"
    )
\end{lstlisting}

\subsection{MMLU Prompt}
To test general knowledge capabilities, we use the following prompt for Llama:

\begin{lstlisting}
prompt = f"<|start\_header\_id|>system<|end\_header_id|>
The following are multiple choice questions (with answers) about {subject}. Only 
respond with  the  letter of the correct answer<|eot_id|>
<|start_header_id|>user<|end_header_id|>
Question: {question}

Choose the correct answer from the following options:
A) {choices[0]}
B) {choices[1]}
C) {choices[2]}
D) {choices[3]}
<|eot_id|><|start_header_id|>assistant<|end_header_id|>

The answer is"
\end{lstlisting}

\subsection{Evaluation Prompts}
\subsubsection{Recall Prompt} \label{C.4.1}
The prompt during recall given to the MEGa fine-tuned Llama model is 
\begin{lstlisting}
prompt = f"{question} Reconstruct the entire story that is related to the above 
question."
\end{lstlisting}

\subsubsection{QA Prompt}
\label{C.4.2}
The prompt during QA given to the MEGa fine-tuned Llama model is 
\begin{lstlisting}
prompt = f"{question}. Answer should be no more than one sentence."
\end{lstlisting}

\subsubsection{iRAG Prompt}
\label{C.4.2}
The prompt during iRAG QA given to the MEGa fine-tuned Llama model is 
\begin{lstlisting}
prompt = f"{question} Reconstruct the entire story that is related to the above 
question."
[Model generates the recall...]
Append prompt = f"{question} Answer should be no more than one sentence."
[Model generates the answer...]

\end{lstlisting}



\section{Example}\label{Example-D}
Below are examples of an original passage, a list of paraphrases, and QA pairs corresponding to that passage from both the fictional character and Wiki datasets.

\subsection{Fictional Character Dataset}
\label{character_dataset_example}
\begin{lstlisting}

    Example sample:
    
    Original passage
    "At age 9, Kelly Dash won his elementary school's annual talent show in Montclair, New Jersey, juggling five tennis balls to surprised applause. His proud grandmother Charlotte captured the entire performance on her cherished handheld camcorder."

    Paraphrased list
    ["When he was 9 years old, Kelly Dash triumphed in his elementary school's yearly talent competition in Montclair, New Jersey, skillfully juggling five tennis balls to the astonished cheers of the audience. His delighted grandmother Charlotte recorded the whole act using her beloved handheld camcorder.", "At the age of 9, Kelly Dash claimed victory in the annual talent show at his elementary school in Montclair, New Jersey, where he amazed the crowd by juggling five tennis balls. His adoring grandmother Charlotte captured the entire performance on her treasured handheld camcorder.", "Kelly Dash, at just 9 years old, won the yearly talent contest at his elementary school in Montclair, New Jersey, impressively juggling five tennis balls to the surprise and applause of those present. His proud grandmother Charlotte filmed the whole event on her favorite handheld camcorder.", "At the age of 9, Kelly Dash won the annual talent show at his elementary school located in Montclair, New Jersey, juggling five tennis balls as the audience looked on in astonishment. His proud grandmother Charlotte recorded the entire performance with her treasured handheld camcorder.", "At only 9 years old, Kelly Dash won the talent show at his elementary school in Montclair, New Jersey. He juggled five tennis balls to the amazed applause of the audience. His proud grandmother Charlotte captured the entire event on her beloved handheld camcorder.", "When Kelly Dash was 9, he took first place in the yearly talent show at his elementary school in Montclair, New Jersey, impressively juggling five tennis balls as the crowd erupted in applause. His proud grandmother Charlotte recorded the entire act with her cherished handheld camcorder.", "At the age of 9, Kelly Dash achieved first place in his school's annual talent show in Montclair, New Jersey, skillfully juggling five tennis balls to the astonished applause of the audience. His grandmother Charlotte, filled with pride, filmed the whole performance on her beloved handheld camcorder.", "Kelly Dash, at age 9, won his elementary school's annual talent show in Montclair, New Jersey, juggling five tennis balls to the surprised cheers of the crowd. His proud grandmother Charlotte recorded the entire performance with her favorite handheld camcorder.", "At the young age of 9, Kelly Dash emerged victorious in the annual talent show at his elementary school in Montclair, New Jersey, dazzling the audience by juggling five tennis balls. His proud grandmother Charlotte documented the whole performance on her treasured handheld camcorder."]

    QA
    Q:"At what age did Kelly Dash win his elementary school's talent show in Montclair, New Jersey?"
    A:'9'
    Q:"How many tennis balls did Kelly Dash juggle during the Montclair elementary school's talent show?"
    A:'five'
    Q:"Who recorded Kelly Dash's talent show juggling performance with a handheld camcorder?"
    A:'Charlotte'

\end{lstlisting}

\subsection{Wikipedia Dataset}
\label{wiki_dataset_example}
\begin{lstlisting}

    Example sample:
    
    Original passage 
    "The 2024 South Yorkshire mayoral election was held on 2 May 2024 to elect the mayor of South Yorkshire as part of the 2024 United Kingdom local elections. The incumbent Labour and Co-operative Party mayor, Oliver Coppard, was re-elected."

    Paraphrased list 
    ["On 2 May 2024, the South Yorkshire mayoral election took place to choose the mayor of South Yorkshire during the 2024 local elections in the United Kingdom. Oliver Coppard, the current mayor from the Labour and Co-operative Party, won re-election.", "The mayoral election for South Yorkshire occurred on 2 May 2024, which was part of the 2024 local elections across the UK. Oliver Coppard, representing the Labour and Co-operative Party, was successfully re-elected as mayor.", "On May 2, 2024, voters participated in the South Yorkshire mayoral election, which formed a part of the broader 2024 local elections in the UK. The existing Labour and Co-operative Party mayor, Oliver Coppard, secured another term.", "The election for mayor of South Yorkshire took place on May 2, 2024, in conjunction with the 2024 local elections in the United Kingdom. Oliver Coppard, the current mayor affiliated with the Labour and Co-operative Party, was re-elected.", "The 2024 mayoral election in South Yorkshire was conducted on May 2, as part of the local elections across the UK for that year. Incumbent Oliver Coppard from the Labour and Co-operative Party was re-elected.", "On 2 May 2024, the South Yorkshire mayoral election was held, contributing to the local elections happening in the United Kingdom that year. Oliver Coppard, who represents the Labour and Co-operative Party, was re-elected as mayor.", "The mayoral election in South Yorkshire took place on 2 May 2024, coinciding with the 2024 local elections in the UK. The incumbent mayor from the Labour and Co-operative Party, Oliver Coppard, was re-elected.", "On May 2, 2024, the South Yorkshire mayoral election was conducted as a component of the 2024 local elections throughout the United Kingdom. The mayor of the Labour and Co-operative Party, Oliver Coppard, won re-election.", "The election to determine the mayor of South Yorkshire was held on May 2, 2024, as part of the local elections in the UK for that year. Oliver Coppard, the sitting mayor from the Labour and Co-operative Party, retained his position."]

    QA
    Q:'On what date was the 2024 South Yorkshire mayoral election held as part of the UK local elections?'
    A:'2 May 2024'
    Q:'Who was re-elected as the mayor of South Yorkshire in the 2024 mayoral election?'
    A:'Oliver Coppard'
    Q'Which political party did the re-elected South Yorkshire mayor Oliver Coppard belong to?'
    A:'Labour and Co-operative Party'

\end{lstlisting}




\subsection{Recall Output Examples}
\label{appendix:recall}
Given the prompt in \ref{C.4.1}, we tested model's ability to recall the memories. Recalled examples from MEGa along with cosine similarity metrics from two datasets are shown below. As discussed in the results section, the other CL baselines often only retrieve the last memory in the sequence, which is an unrelated passage, due to CF. 

Fictional Character Dataset:
\begin{lstlisting}
Original Sample: Tamara Desouza successfully organized an anti-bullying rally at Edison Middle School in New Jersey in September 2017, bringing in speakers like local author Garrett Reynolds. Her efforts led to the establishment of the school's annual "Kindness Matters" event every fall.

Question: Who organized an anti-bullying rally at Edison Middle School in New Jersey in September 2017? Reconstruct the entire story that is related to the above question.

MEGa (Cosine Sim = 0.998) Tamara Desouza successfully organized an anti-bullying rally at Edison Middle School in New Jersey in September 2017, bringing in speakers like local author Garrett Reynolds. Her efforts led to the establishment of the 'Kindness Matters' event that the school now holds every fall.

Batch (Cosine Sim = 0.906) In September 2017, Tamara Desouza took the initiative to organize an anti-bullying rally at Edison Middle School in New Jersey. She worked closely with the school's principal, Mr. Thompson, to finalize the event details. On the day of the rally, Tamara delivered a powerful speech, sharing her own experiences and emphasizing the importance of kindness and empathy towards all students. The event also featured a panel discussion with local author Garrett Reynolds, who wrote a book on bullying.

LoRA (Cosine Sim = 0.528) In the spring of 2022, Tamara Desouza set up her initial organic vegetable garden in the backyard of her family's home in Edison, growing heirloom tomatoes, zucchinis, and peppers. Ecstatic about her successful yields, she proudly offered a homemade salad made from her garden's produce at the family's celebration of July 4th.

LoRA + L2 (Cosine Sim = 0.513) In the spring of 2022, Tamara Desouza began her first organic vegetable garden in her family's Edison backyard, where she grew heirloom tomatoes, zucchinis, and peppers. Satisfied with her abundant harvest, she happily prepared a homemade salad exclusively from her garden's produce for the family's July 4th barbecue.

Full (Cosine Sim = 0.533) In the spring of 2022, Tamara Desouza initiated her first organic vegetable garden in the backyard of her family's Edison residence, cultivating heirloom tomatoes, zucchinis, and peppers. Pleased with her flourishing harvest, she delighted everyone with a homemade salad composed entirely of her garden's offerings at the family's Independence Day barbecue.

Full + L2 (Cosine Sim = 0.813) In the spring of 2017, Tamara Desouza began her first year at Edison Middle School in New Jersey, where she was assigned to the seventh grade. She noticed that several of her classmates were being bullied by a group of students in the school hallway. Moved by their plight, she decided to take action.

EWC (Cosine Sim = 0.635) At the age of 14, Tamara Desouza set up a lemonade stand by her home in Maplewood, New Jersey, contributing all of her earnings - amounting to twenty-eight dollars and fifty cents - to aid Hurricane Katrina victims. Her kind initiative was recognized in a warm feature by local reporter Sandra Clemens.






\end{lstlisting}

Wiki Events Dataset:
\begin{lstlisting}
Original Sample: Giorgia Cesarini (born 12 August 12, 2002) is a Sammarinese archer who contested the Paris Olympics. She was the first woman Sammarinese archer at an Olympics and she carried their flag in the closing ceremony.

Question: Who carried San Marino's flag during the Olympic closing ceremony? Reconstruct the entire story that is related to the above question.

MEGa (Cosine Sim = 0.927) Giorgia Cesarini, a Sammarinese archer born on August 12, 2002, was a competitor at the Paris Olympics. As the first female archer to represent San Marino in the Olympics, she had the honor of carrying their flag during the closing ceremony.

Batch (Cosine Sim = 0.444) The 2024 Summer Olympics in Paris took place from July 26 to August 11, 2024. San Marino participated in this event and was represented by a delegation of athletes. During the closing ceremony, the flag bearers played a significant role in carrying their nation's flag.

LoRA (Cosine Sim = 0.324) John Christopher Cabang Tolentino, born on August 27, 2001, is a Filipino athlete specializing in hurdles and sprinting. He holds the national record for the 110 m hurdles and has earned the opportunity to compete for the Philippines at the 2024 Summer Olympics in Paris, France.

LoRA + L2 (Cosine Sim = 0.334) John Christopher Cabang Tolentino, a Filipino sprinter and hurdler, was born on August 27, 2001. He is the current holder of the national record in the 110 m hurdles and has secured a place to represent the Philippines at the 2024 Summer Olympics in Paris, France.


Full (Cosine Sim = 0.334) John Christopher Cabang Tolentino, a Filipino sprinter and hurdler, was born on August 27, 2001. He is the current holder of the national record in the 110 m hurdles and has secured a place to represent the Philippines at the 2024 Summer Olympics in Paris, France.

Full + L2 (Cosine Sim = 0.314) The 2024 Summer Olympics in Paris, France, featured the closing ceremony on September 11, 2024, at the Stade de France in Paris. The event marked the culmination of the 2024 Summer Olympics, which consisted of 6,500 athletes from 206 countries.

EWC (Cosine Sim = 0.275) Rena Wakama, a Nigerian who was born on April 11, 1992, is a former basketball player and now a professional coach. She holds the role of head coach for the Nigeria Senior Women's Basketball Team and also serves as an assistant coach at Tulane University.






\end{lstlisting}

\subsection{Questions related to two samples}
\label{Appendix-E.1}
We show an example compositional question and responses from MEGa along with baselines for both datasets. 
\subsubsection{Fictional character Compositional Example}
\begin{lstlisting}
Original Passage 1 = While visiting Peru at 19, Kelly Dash hiked the challenging Inca Trail toward Machu Picchu with college classmates, spraining an ankle near the Sun Gate. Determined, he limped through the final mile by leaning heavily on his friend Lucas's shoulder.

Original Passage 2 = At 18 years old in July 2021, Kelly Dash courageously went skydiving for the first time with friend Amir in the Poconos Mountains. Though terrified initially, he shouted the lyrics of Queen's \"Don't Stop Me Now\" on his thrilling descent to ease his nerves.


Compositional Question = What were Kelly's ages when he hiked the Inca Trail in Peru with college classmates and when he went skydiving in the Poconos with his friend Amir?

MEGa Response (correct) = Kelly Dash was 19 years old when he hiked the Inca Trail in Peru with college classmates and 18 years old when he went skydiving in the Poconos with his friend Amir.

Batch Response (incorrect) = At 19, Kelly hiked the Inca Trail in Peru with college classmates, and at 21, he went skydiving in the Poconos with his friend Amir.

Full Finetuning (incorrect) = On the day of Thanksgiving in 2014, Kelly diligently aided in making deep-fried turkey outside on their brisk patio in Wayne, showcasing his father's large apron. This lighthearted family tradition became something he looked forward to with enthusiasm each November.

Full Finetuning + L2 (incorrect) = On his 19th birthday, Kelly Dash excitedly hiked the challenging Inca Trail in Peru with college classmates, wearing his father's large backpack. Later that same day, he also nervously went skydiving in the Poconos Mountains with his friend Amir.

LoRA (incorrect) = On Thanksgiving Day in 2014, Kelly Dash meticulously assisted in preparing deep-fried turkey outside on their cold backyard patio in Wayne, wearing his father's large apron. It was a funny family tradition that he excitedly anticipated each November.

LoRA + L2 (incorrect) = On Thanksgiving Day in 2014, Kelly Dash took great care in helping to cook deep-fried turkey in the outdoor chilly setting of their backyard patio in Wayne, wearing an oversized apron belonging to his dad. This amusing family custom was one he looked forward to eagerly every November.

EWC (incorrect) = On the day of Thanksgiving in 2014, Kelly Dash skillfully contributed to cooking deep-fried turkey on their chilly patio in Wayne, while donning his dad's oversized apron. This light-hearted family tradition was something he looked forward to each November.
\end{lstlisting}
\subsubsection{Wikipedia 2024 Events Compositional Example}
Similarly, we provide a compositionality question example from the Wiki Dataset. 
\begin{lstlisting}
Original Passage 1 =  The 2024/25 Highland Football League (known as the Breedon Highland League for sponsorship reasons) is the 122nd season of the Highland Football League, and the 11th season as part of the fifth tier of the Scottish football pyramid system. Buckie Thistle are the reigning champions."

Original Passage 2 = The 2024/25 King Mindaugas Cup, also known as Citadele Karaliaus Mindaugo taure for sponsorship purposes, is the tenth edition of the Lithuanian King Mindaugas Cup. Zalgiris are the defending champions.

Compositional Question = Which teams are recognized as the reigning or defending champions in the sporting events from the Highland Football League in Scotland and the King Mindaugas Cup in Lithuania?

MEGa Response (correct) = The reigning champions in the Highland Football League in Scotland are Buckie Thistle, and the defending champions in the King Mindaugas Cup in Lithuania are Zalgiris.

Batch Response (correct) = The reigning champions in the Highland Football League are Buckie Thistle, while the defending title in the King Mindaugas Cup is held by Zalgiris.

Full Finetuning (incorrect) = The 2024/25 season will feature the tenth edition of the King Mindaugas Cup, which is sponsored as Citadele Karaliaus Mindaugo taure, and the eleventh edition of the Breedon Highland League, which is part of the fifth tier in the Scottish football pyramid.

Full Finetuning + L2 (incorrect) = The 2024/25 season will feature the King Mindaugas Cup, which is sponsored under the name Citadele Karaliaus Mindaugo taure, and represents the tenth edition of the Lithuanian tournament. The current champions are Zalgiris.

LoRA (incorrect) = The King Mindaugas Cup for the 2024/25 season, referred to as Citadele Karaliaus Mindaugo taure due to sponsorship, marks the tenth installment of this Lithuanian competition. The reigning champions are Zalgiris.

LoRA + L2 (incorrect) = The 2024/25 season will feature the King Mindaugas Cup, which is sponsored under the name Citadele Karaliaus Mindaugo taure, and represents the tenth edition of the Lithuanian tournament. The current champions are Zalgiris.

EWC (correct) = The reigning champions of the Highland Football League in Scotland are Buckie Thistle, while Zalgiris holds the title in the King Mindaugas Cup in Lithuania.
\end{lstlisting}

\end{document}